\documentclass[letterpaper, 10 pt, conference]{ieeeconf}  
\IEEEoverridecommandlockouts               

\usepackage{times}
\usepackage{epsfig}
\usepackage{graphicx}
\usepackage{float}
\usepackage{wrapfig}
\usepackage{amsmath,amssymb}
\usepackage{bm,xspace}
\usepackage{comment}
\usepackage{verbatim}
\usepackage{multirow}
\usepackage{balance}
\usepackage{url}
\usepackage{booktabs}
\usepackage{etoolbox,siunitx}
\usepackage{calc}
\usepackage{pifont,hologo}
\usepackage{graphicx}
\usepackage{xcolor}

\usepackage{multicol}
\usepackage{algorithm2e}
\usepackage{makecell}
\usepackage{arydshln}
\usepackage{multirow}
\usepackage{ulem}

\usepackage{thmtools}
\usepackage{thm-restate}

\newcommand{\ra}[1]{\renewcommand{\arraystretch}{#1}}

\usepackage[super]{nth}
\usepackage{nicefrac}
\robustify\bfseries

\def\dd{\mathbf{d}}

\def\pp{\mathbf{p}}

\def\sss{\mathbf{s}}
\def\tt{\mathbf{t}}

\def\vv{\mathbf{v}}

\def\DD{\mathbf{D}}

\def\II{\mathbf{I}}

\def\LL{\mathbf{L}}
\def\MM{\mathbf{M}}

\def\OO{\mathbf{O}}

\def\SS{\mathbf{S}}

\def\dD{\mathcal{D}}

\def\gG{\mathcal{G}}

\def\nN{\mathcal{N}}

\def\xX{\mathcal{X}}

\DeclareMathOperator*{\argmin}{arg\,min}

\DeclareMathSymbol{@}{\mathord}{letters}{"3B}

\newcommand\norm[1]{\left\lVert#1\right\rVert}

\def\latex/{\LaTeX}
\def\bibtex/{\hologo{BibTeX}}

\usepackage{adjustbox}

\title{\LARGE \bf
OF-VO: Efficient Navigation among Pedestrians Using Commodity Sensors
}

\author{
Jing Liang, Yi-Ling Qiao, Tianrui Guan and Dinesh Manocha 
}

\begin{document}
\maketitle
\thispagestyle{empty}
\pagestyle{empty}

\begin{abstract}
We present a modified velocity-obstacle (VO) algorithm that uses probabilistic partial observations of the environment to compute velocities and navigate a robot to a target. Our system uses commodity visual sensors, including a mono-camera and a 2D Lidar, to explicitly predict the velocities and positions of surrounding obstacles through optical flow estimation, object detection, and sensor fusion.
A key aspect of our work is coupling the perception (OF: optical flow) and planning (VO) components for reliable navigation. Overall, our OF-VO algorithm using learning-based perception and model-based planning methods offers better performance than prior algorithms in terms of navigation time and success rate of collision avoidance. Our method also provides bounds on the probabilistic collision avoidance algorithm. We highlight the realtime performance of OF-VO on a Turtlebot navigating among pedestrians in both simulated and real-world scenes. A demo video is available at \url{https://gamma.umd.edu/ofvo/}

\end{abstract}

\linespread{0.93}
\section{INTRODUCTION}
% introduce robots utilities and prosperities
% introduce centralised/decentralised ways.
% introduce different dynamic based approaches
% introduce our contributions and description of our approach
% introudce and compare learning based algorithms
% introduce the CV algorithms for segmentation and velocity extraction

Mobile robots are currently deployed in a wide range of scenarios, including warehouses, airports, malls, and offices ~\cite{ma2017overview}. In these indoor and outdoor spaces, robots are used to perform routine tasks such as delivering goods or guiding customers. The ability to perform reliable navigation and avoid collisions with pedestrians and other obstacles is  important to perform these tasks.
%Safely interacting with humans and avoiding collisions is a mobile robot's priority when moving in the crowd.

%However, collision avoidance algorithms for robot navigation still remain far from perfect as a result of the variety of possible environments and limitations on robot perception. 
There are extensive works on robot navigation and collision-avoidance in dynamic scenes~\cite{dwa, VO}. These include model-based techniques based on velocity obstacles~\cite{orca} and dynamic constraints~\cite{dwa}. However, these methods assume accurate detection or representation of obstacles in the environment \cite{orca, dwa}. %AMRCOMMENT: I tend to dislike using "etc" but others may disagree
% Many extensions have been proposed for real-world  scenes using vector-field-based approaches ~\cite{babinec2014vfh,shimoda2005potential}, which only work well for static obstacles. 
Other techniques have been proposed for dynamic scenes~\cite{VO,levy2015extended}, but their performances vary in different environments and none of them are robust enough for all general scenarios~\cite{li2018role}. 
% Furthermore, some methods like ~\cite{pfrunder2017real} rely on very expensive sensors like RTK-GPS or 3D Lidars.

%efficacy\cite{li2018role}. Velocity obstacle \cite{orca, hennes2012multi, orca-dd, nh-orca} approaches can provide safe strategies for robots to avoid collisions but the limitation is the assumption that robots need to have accurate state information, including relative position and velocity, of nearby robots. In this project, we propose a novel architecture to solve this issue and get accurate observations of nearby obstacles with sensors of RGBD camera and lidar.

%~\cite{babinec2014vfh,levy2015extended,fiorini1998motion,shimoda2005potential}, which work well in mostly static environments or use expensive visual sensors (e.g., 3d lidars).

Recently, a number of learning-based approaches have been proposed to perform robot navigation and collision avoidance~\cite{sathyamoorthy2020densecavoid,CesarCadena,Target_driven}. These algorithms can handle sensor noise in many scenarios. 
%and tend to recover the expert policy in many scenarios. 
However, it is challenging to predict the performance in new or unknown scenes with those learning-based models because their performance is governed by the training data.  
%There is no guarantee that we can gain the same performance in all scenarios. 
%Therefore, explainability and interpretability are essential aspects of a good navigation algorithm. 
Some model-based navigation algorithms~\cite{PRVO, park2017efficient} use probabilistic techniques to handle sensor errors, but they are not robust in terms of handling partial observations. 

\begin{figure}
\centering
\begin{tabular}{c@{\hspace{1mm}}@{\hspace{1mm}}c}
    \includegraphics[width=.45\linewidth, height=.3\linewidth]{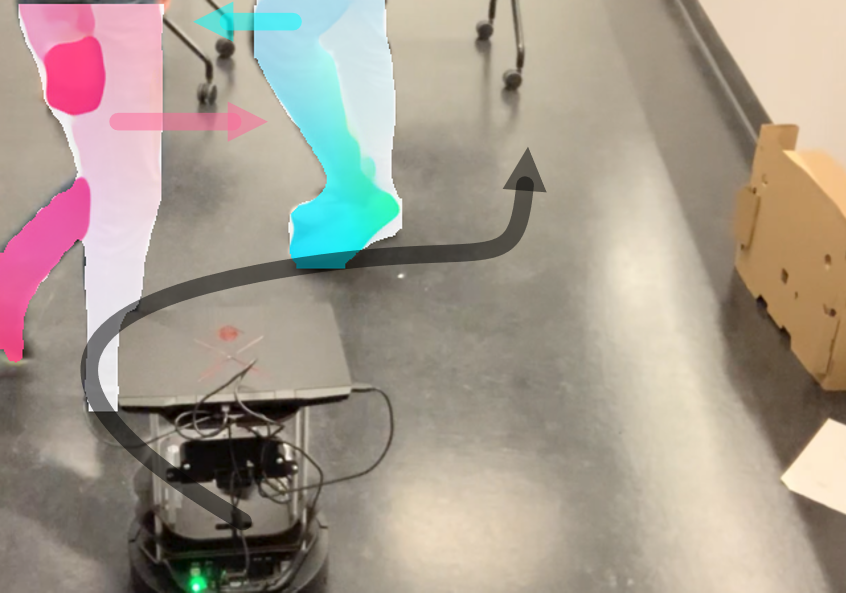}&
    \includegraphics[width=.45\linewidth, height=.3\linewidth]{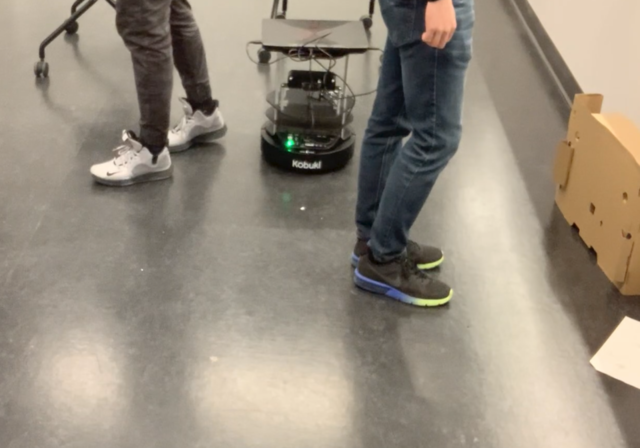} 
    % \small (a) segmentation & \small (b) optical flow + mask
\end{tabular}
\caption{Navigation among pedestrians in an indoor scene: We highlight the scene captured using a commodity camera mounted on the robot. The image on the left shows the segmentation of pedestrians with colored optical flow computed by our perception algorithm. Blue and red arrows correspond to pedestrian velocities. The curved black arrow  represents the robot's trajectory computed using our probabilistic collision avoidance algorithm. The image on the right highlights the environment in which the robot navigates among the pedestrians. }
\vspace*{-4mm}

% The top row highlights the segmentation and optical flow computed using our algorithm and the straight red arrow highlights the computed pedestrian velocity.
% Based on the perception, in bottom left figure the curved black arrow represents the trajectory of robot, and blue and red arrows are velocities of pedestrians. Top right picture shows the result state of the collision avoidance. 

% red and black arrows represent the trajectories of the pedestrian and the robot, respectively. The bottom row highlights the segmentation and optical flow computed using our algorithm and the straight red arrow highlights the computed pedestrian velocity. and we use this information for collision-free trajectory computation. 
\label{fig:teaser}
% \vspace{-10mm}
\end{figure}
% \vspace*{0.1in}

%AMRCOMMENT: you mention interpretability and explainability here but you don't discuss it ever again
\noindent {\bf Main Results:} We present a novel hybrid scheme for safe navigation of a mobile robot among pedestrians and other obstacles. Our approach takes advantage of learning-based methods to handle sensor data (perception) and a model-based formulation (planning) to compute safe velocities for collision avoidance. Instead of developing an end-to-end learning pipeline, we use separate components for perception and planning to provide guarantees of collision avoidance. The novel components of our approach include:
\begin{itemize}
   % \item We close the sim-to-real gap by using a high-fidelity simulator with real-world crowd behaviors, sensor noises, and robot constraints. We test the framework in multiple standard scenarios and more complex environments to make it generalized. The new framework supports expansion for tasks with different purposes.
    \item We integrate a segmentation and an optical flow estimation network to detect obstacles. The state-of-the-art pretrained models helps robots perceive the environment accurately in real time.
    % \item We develope method by using pretrained segmentation and optical flow network to detect static and dynamic obstacles. The method does not need additional fine-tuning and is fast and generalizable to real-world data.
    \item We present a modified velocity-obstacle algorithm to perform local navigation and collision avoidance based on partial observations. We highlight improvements over prior velocity obstacle methods and show that our approach is more robust.
    \item We present bounds for our probabilistic collision avoidance. These bounds provide better guarantee and interpretability than learning-based navigation schemes.
    % Based on these bounds, we choose the time intervals to provide the desired confidence of collision avoidance.
    %We present a modified velocity obstacle algorithm that uses partial observations corresponding to optical flows as inputs and computes a collision-free trajectory for the robot.
  %  \item We demonstrate improved performance over prior methods based on dynamic windows and deep reinforcement learning. 
    %better performance over alternatives in terms of success rate, trajectory length, and navigation time.
    %We present a real-world navigation scheme (OF-VO) that uses commodity sensors corresponding to RGB-D cameras and 2D lidars that can be easily mounted on mobile robots.
\end{itemize}

We have tested our algorithm on a Turtlebot robot 
 with commodity sensors, including an Orbecc Astra RGB camera (around \$100) and a 2-D Hokuyo Lidar (costing around \$1000). 
 We have evaluated OF-VO in simulated environments with multiple pedestrians, including high-density cases with more than $10$ pedestrians in $5 \times 5 \; m^2$ area. 
 We used different methods to simulate pedestrian movements, including socially acceptable collision avoidance~\cite{shiomi2014towards,bera2017sociosense}. 
Moreover, we tested the algorithm in our laboratory with pedestrians close to the robot and moving in different patterns. Results will be shown in the video. %AMRCOMMENT:  add a \space after 5\times 5
 We compared our approach with prior model-based methods like DWA~\cite{dwa} and deep-reinforcement learning algorithms like~\cite{fan2018fully}. We observe $2-5X$ improvement in success rate (and reliability).
%  and around $40\%$ reduction in the navigation time in complex scenarios.

 %compared to alternative model-based and learning-based methods. 
\section{BACKGROUND AND RELATED WORK}
In this section, we briefly survey related work on local navigation and collision avoidance. We also give an overview of prior techniques used in computer vision for object segmentation and motion detection.

\subsection{Collision Avoidance}
 Many researchers have proposed learning-based algorithms for collision avoidance in real-world scenarios. \cite{POMDP} uses POMDPs to navigate robots in uncertain scenarios. \cite{CesarCadena} and \cite{Target_driven} use deep learning to train end-to-end networks for indoor scenarios. These methods have shown good performance in terms of avoiding static obstacles. However, these methods may not work well in complex dynamic scenes, especially when the pedestrians are moving at different speeds. In addition to considering the uncertainties of perception module, our approach also takes into account different velocities of pedestrians.

Many model-based algorithms~\cite{orca,dwa,wolinski2014parameter} have been proposed for motion planning in dynamic scenarios. However, these methods assume accurate detection, localization  or representation of obstacles. In real-world scenarios, perception methods tend to be inaccurate due to sensor noise and environment complexity~\cite{kim2015brvo}. To address the inaccuracies, \cite{forootaninia2017uncertainty} proposes a time-to-collision algorithm for reciprocal collision avoidance, though it is mainly limited to simulated scenarios.  \cite{ du2011robot} uses a receding horizon control algorithm with different constraints to navigate robots in uncertain environments. However, this approach can be time-consuming as the algorithm needs to compute several consecutive actions in one step. \cite{PRVO, kim2015brvo, angeris2019fast} use velocity obstacles for collision avoidance. However, these methods assume complete observations of nearby obstacles and environments. 
% \cite{ARXIV}. JUST CITE THE ARXIV PAPER, AND NO NEED TO MENTION THE APPENDIX SEPARATELY. 
%the appendix\footnote{Appendix: \url{https://arxiv.org/abs/2004.10976}}
% ARXIV FIGURE. COPY THE FIGURE IN COVER LETTER TO THE ARXIV PAPER AND CITE THAT HERE. 
We also take into account the kinematic constraint of robot and apply the constraint in the  velocity space.
% KINEMATIC CONSTRAINTS OF WHAT (THE ROBOT OR OBSTACLE) 
% with partial observations of the nearby obstacles.  
Overall, our approach results in improved navigation behavior.

Neural networks techniques have been used for navigation with real-world sensors.  For example, cameras and Lidar are widely used as inputs to networks in reinforcement learning approaches. For RGB cameras, \cite{D3QN} uses deep double-Q and \cite{mnih2016asynchronous} uses the A3C algorithm to train the policies in cluttered scenarios.
%, and \cite{Target_driven} uses the Target-final algorithm to train policies based on RGB images in indoor scenarios. 
\cite{fan2018fully} uses 2D Lidar along with the PPO algorithm~\cite{schulman2017proximal} to navigate multiple robots in scenarios with static and dynamic obstacles. Some deep reinforcement learning-based navigation algorithms use a combination of cameras and 2D Lidar~\cite{sathyamoorthy2020densecavoid}. 
\cite{henry2010learning} uses inverse reinforcement learning to imitate the behavior of people in crowds. However, these methods can result in oscillation or freezing issues during navigation. Furthermore, it is difficult for these learning-based approaches to provide strong guarantees on collision avoidance in general or arbitrary scenarios. In contrast, our approach only uses learning based methods in the perception module and can provide proabilistic bounds on collision avoidance and navigation.

\subsection{Motion Estimation}
Estimating the positions and velocities of moving objects plays an important role in perception tasks. Previous works estimate velocities through trajectory tracking and prediction~\cite{henschel2018fusion,sathyamoorthy2020densecavoid}. 
% However, tracking algorithms may not perform consistently well in dense scenes. They may lose tracking targets~\cite{li2019estimation}, which may result in inaccurate position and velocity estimations. 
Tracking methods can exhibit higher accuracy when the pedestrian is fully contained in the field-of-view of the cameras. However, when a moving pedestrian is very close to the robot, it may be only partially visible, and the tracking algorithms might fail due to the loss of tracking or mismatch across frames~\cite{li2019estimation}. Instead, we use an optical flow network, which is more robust and the network encodes the movement of objects in pixel-level correspondence.
To extract the velocities of objects, our perception approach consists of two modules: optical flow estimation and object detection (see Fig. 2). 

Optical flow (OF) is the displacement of the pixel position between two consecutive images. Sparse OF only computes the flow on some landmark points, which takes less time. For real-time performance, other methods~\cite{tchernykh2006optical,chao2014survey} use sparse OF to avoid collisions. They use the property that pixels from closer objects tend to move faster, and thus the robot steers to the side with smaller optical flow. Although this rule seems direct and intuitive, the underlying assumption may not hold in complex dynamics scenes. Recently, \cite{rashed2019optical} augments segmentation using optical flow estimation. \cite{Honegger2012real} estimates velocity based on optical flow estimation, but does not perform any segmentation computations. Our estimation using dense segmentation and optical flow is more accurate than prior methods. Traditional dense optical flow estimation can be achieved by optimization with smoothness constraints~\cite{horn1981determining}, which takes several minutes to compute on a single image pair. Some learning-based methods based on neural networks~\cite{dosovitskiy2015flownet,ilg2017flownet,sun2018models,sun2018models} have also been designed to estimate the optical flow with accuracy comparable to traditional methods, while running an order of magnitude faster. However, they do not provide accurate average velocity for each moving object, given the fact that different parts of one object may have different velocities. 
% In our method, we combine FlowNet2~\cite{ilg2017flownet} and m to estimate the motion between two frames.
In order to achieve a high accuracy in velocities prediction, we use the state-of-the-art models MaskRCNN~\cite{he2017mask} and FlowNet2~\cite{ilg2017flownet} to calculate the average speed of objects. Our pipeline does not restrict the choices of networks and our current implementation uses these two networks because they are widely used in computer vision. MaskRCNN is implemented in most open-source object detection frameworks~\cite{wu2019detectron2,mmdetection}. FlowNet2 is one of the most popular methods for optical flow estimation. These two networks are trained on a large volume of data. We use them in simulated environments as well as real-world scenarios.

% State-of-the-art detection methods like~\cite{girshick14CVPR} and \cite{renNIPS15fasterrcnn} can detect different categories of objects and compute their bounding boxes. These learning-based methods can predict with high efficiency and accuracy. In order to estimate the motion with dense optical flows, we need to segment the objects from the background. Mask-RCNN~\cite{he2017mask} and YOLO~\cite{redmon2017yolo9000} are able to compute pixel-level segmentation of different objects.

Lidar sensors can exhibit good performance in terms of computing distances~\cite{wei2018lidar}, and cameras are widely used for detection estimation~\cite{redmon2017yolo9000}. The combination of these two sensors has been used for navigation and autonomous driving~\cite{debeunne2020review}. 
In our method, we fuse the 2D data from Lidar and the  images from cameras to compute the obstacles' relative positions and improve the velocity prediction.
% Many works fuse data from Lidar and cameras \cite{kumar2020lidar,ramezani2018end}. In our method, we combine 2D data from Lidar with detected objects from images to compute the obstacles' relative positions and velocities.
% \jing{ADD SENSOR FUSION}

%\subsection{Sim-to-Real and Behavioral Constraints}

%Simulations can provide abundant data and are not bound to the safety and space constraints, but sometimes they are not ideal due to modeling errors. There are many efforts on closing the sim-to-real gap \cite{crowdsteer, sim2real, biped}, either to provide realistic simulations and add more constraints on the physiological and psychological factors \cite{braun, narain, narang, pelecha}, or to minimize the noise from the sensor \cite{noisysensor}, so that the original performance of the algorithm in the simulator would remain at implementation time.

\section{Overview}
\label{section:overview}

\begin{figure*}[t]
\vspace{1mm}
  \centering
  \includegraphics[width =.92\linewidth,height=.28\linewidth]{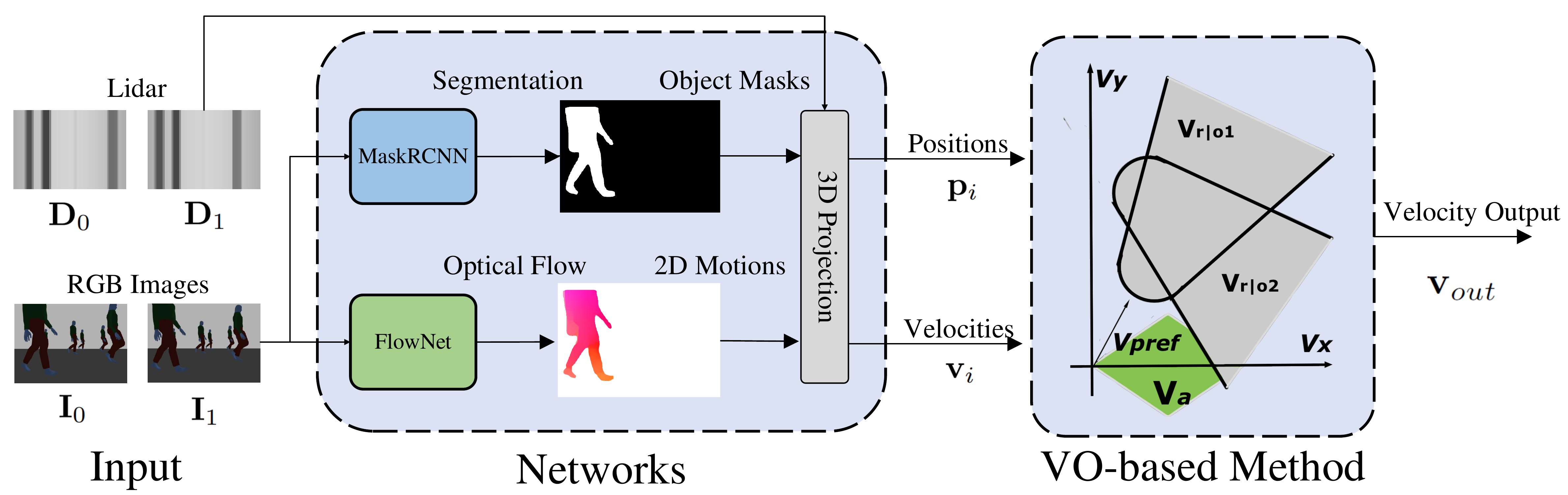}
  \caption {OF-VO Pipeline: During each time step, our method first takes as input a pair of RGB images and Lidar sequences. The optical flow and segmentation masks are computed using FlowNet and MaskRCNN, respectively. Using Lidar data and intrinsic camera parameters, our 3D projection module computes the positions and velocities from the masks and optical flow. Due to the limitation of the camera's FOV, 2D Lidar data is also collected to enhance the position estimation. Our method uses a modified velocity obstacles approach to compute a collision-free velocity for the robot that accounts for sensor errors. Before sending the control signals to the robot, a differential driving formulation is used to satisfy the kinematic constraints of the robot.  
  %Our method takes advantage of neural networks pretained on large real-world dataset, therefore, avoid excessive training, finetuning, and sim-to-real. The networks and VO are collaborative and have strong coupling. The learning-based perception module can provide accurate up-to-date velocities for the following VO-based algorithm. In turn, the partially observable VO can be set as more conservative to mitigate the error in perception.
  }
  \label{fig:arch}
\vspace*{-5mm}
\end{figure*}

% obstacle velocity: $\vv_i$
% obstacle position: $\pp_i$
% preferred velocity: $\vv_{pref}$
% preferred velocity: $\vv_{out}$
% $V_{feasible}$

Our approach is designed for non-holonomic robots, whose movement is governed by linear and angular velocities. During each timestep, the robot knows the relative position of its target/goal and needs to navigate to the target in an environment that consists of static and dynamic obstacles. Each obstacle has a velocity $\vv_{i}$ at position $\pp_{i}$.  We do not assume that the dynamic obstacles like the pedestrians would completely avoid the robot and design an appropriate navigation scheme. Due to sensor errors, the velocity and the position values may have some uncertainty.
%AMRCOMMENT: "all the perceived information may have inaccuracies" does this just mean "there might be sensor error" or something else?  a bit awkward phrasing if so
%Dynamic obstacles typically correspond to moving pedestrians. 
In each time step, the robot observes two consecutive RGB and Lidar frames. Images are used by MaskRCNN and FlowNet and then fused with Lidar data to predict the velocities and positions of the moving objects (usually pedestrians) in the scene. The predicted positions and velocities are used in our probabilistic collision avoidance algorithm for local navigation.
% On the way to the goal, the robot can perceive surrounding obstacles by using a 2D Lidar and an RGB camera to avoid potential collisions. 
Figure~\ref{fig:arch} shows the workfolw of OF-VO. 

We compute the velocities and positions of surrounding obstacles from  Lidar data and RGB images with the help of optical flow estimation and object detection. After that, noises from the perception models are explicitly analyzed and used to improve the overall accuracy of our approach.
With the position and velocities, $\pp_o,\; \vv_o$, of detected obstacles, our method uses a modified VO-based algorithm to navigate the robot. 

Real-world scenarios inevitably have inaccuracies, occlusions, and only partial observations from the environment due to limited field-of-view. 
% , a robot can only obtain partial observations from the environment. 
% Different from traditional VO algorithms where the states of the obstacles are fully observable, our OF-VO is able to generate collision-free paths under partial observations. We can provide guarantees on its performance if certain assumptions are satisfied. 
For all obstacles detected by the perception model, we model the inaccuracies in detection as
Gaussian distributions. We also provide a lower bound on the confidence of collision avoidance.
For unobserved or partially observed objects, we apply non-holonomic kinematic constraints to estimate the feasible velocity space of the robot, as described in Section \ref{sec:navigation}. 
Our integrated approach consists of two parts, learning-based perception and model-based collision avoidance, which are strongly coupled.
The hyperparameters in the VO algorithm are chosen according to different bounds on robots' velocities and different requirements in terms of confidence in probabilistic collision avoidance. 
%Under all those constraints, robot can efficiently navigate in uncertain environment.
% obstacles under partial observed conditions, we applied 
% With unobserved objects of states ${\pp'_j,\vv'_j}$, OF-VO applies several reasonable constraints $\CC(\pp'_j,\vv'_j,\ww)$ on the non-visible objects, where $\ww$ represents the hyper-parameters of these constraints. The partial VO algorithm can output a velocity $\vv_{out}=VO(\pp_o,\vv_o,\ww)$ based on the observation. In the end, we use differential drive formulation to convert $\vv_{out}$ to control signals that satisfy the dynamics constraints of the robot. 

%thereby improving the reliability and interpretability of our approach. 
%Moreover, these two modules are strongly coupled. 

%The perception part can efficiently detect and estimate the states of the obstacles, especially the ones closest to the robot.

%and provide up-to-date states to the VO, which can benefit VO to successfully avoid nearby obstacles.
% As depicted in Sec.~\ref{sec:navigation}, with the state of observable obstacles $\pp_o, \vv_o$ 

%Given The navigation method, which is , would generate a velocity in the Euclidean coordinate, $\vv_{VO}$. We propose a novel method to convert the Euclidean velocity to a Polar velocity $\vv_{out}$ for differential drive. Using the $\vv_{out}$, the robot safely avoids collisions in timestep T and finally goes to the goal safely.

\section{PERCEPTION}\label{sec:perception}
In our approach, we use two commodity visual sensors: an RGB camera and a 2D Hokuyo Lidar. We choose to use these two commodity sensors because the RGB camera is good at perceiving detailed information in the scene, while a 2D Lidar can accurately detect distances of nearby objects.
The input to our perception module is a pair of consecutive RGB frames and Lidar data, and the output is the estimated positions $\pp_o$ and velocities $\vv_o$ of the nearby obstacles.
Our algorithm builds on recent learning-based computer vision methods. We use an optical flow network FlowNet~\cite{ilg2017flownet} and a segmentation network MaskRCNN~\cite{he2017mask} to perceive the objects' velocities and positions. The input to the neural networks is a pair of images, and we compute the positions and velocities of objects appearing in the images.

At each time step, we need two consecutive frames to calculate the instantaneous velocities. Assume that the current time step is $t_1$ and the previous time step is $t_0<t_1$. From the RGB camera we have two pairs of RGB images $\II_{t_1},\II_{t_0}\in\mathbb{R}^{h\times w\times 3}$, where $h,w$ is the height and width of the images in pixels. For simplicity, we omit $t$ in the subscripts.
The input to the FlowNet is the RGB pair $\II_1,\II_0$, and the output is the optical flow $\OO\in\mathbb{R}^{h\times w\times 2}$, which represents the displacement of pixels in the 2D images. Therefore, one pixel $\sss_1=[x_1,y_1]$ in $\II_1$ corresponds to $\sss_0=[x_0,y_0]=[x_1+\OO(x_1,y_1,1)),y_1+\OO(x_1,y_1,2)]$ in $\II_0$. Assume the Lidar sequence is $\LL=[l_1,l_2,...,l_k]$, where $l_i$ is the distance to the nearest point at angle $\theta_i$. The positions of those points in the horizontal plane can then be computed using $(l\sin\theta, l\cos \theta)$ and projected into the image space as a depth image $\DD$. The depth $\DD_0'$ and position $\SS_0'$ of pixels in the $\II_1$ frame at $t_0$ time can be computed using optical flow warping~\cite{ilg2017flownet}.

% To compute the displacement in the 3D world, we need to transform the pixel coordinates to camera coordinates with the help of the Lidar. For a camera whose principal point is $[c_x,c_y]$ and focal lengths are $(f_x,f_y)$, the 3D displacement $\sss=[\sss_x,\sss_y,\sss_z]\in\mathbb{R}^{h\times w\times 3}$ from $\pp_1$ to $\pp_0$ can be computed by,
% \begin{align}
%     s_x&=(x_0\cdot z_0-c_x*z_0-x_1\cdot z_1+c_x*z_1)/f_x,\notag  \\
%     s_y&=(y_0\cdot z_0-c_y*z_0-y_1\cdot z_1+c_y*z_1)/f_y,\notag \\
%     s_z&=z_0-z_1.
% \label{eq:3dflow}
% \end{align}
% Finally, the 3D velocity of each pixel in the image $\II_1$ is given as $\vv=\sss/(t_0-t_1)$. Note that $\vv\in\mathbb{R}^{h\times w\times 3}$ is a velocity field defined on the image.

The next step is to separate moving obstacles such as pedestrians from the raw image so that we can compute their positions and velocities. 
We feed image $\II_1$ to another segmentation network MaskRCNN to estimate pixel-level segmentation. Assume that for an object $i$, MaskRCNN predicts its segmentation mask $\MM_i\in\{0,1\}^{h\times w}$. The depth $d_i$ and the position $\sss_{i}=(x_{i}, y_{i})$ in the image space of an object can be calculated as the weighted mean in the masked area.  
%$d_i=(\sum\odot \MM_i)/(\sum\MM_i)$, $\pp_i=(\sum\vv\odot \MM_i)/(\sum\MM_i)$ where $\odot$ is the point-wise product. 
Let the camera have principal point $[c_x,c_y]$ and focal lengths $(f_x,f_y)$. The 3D position of the object can be computed as
$\pp_i=[(x_i\cdot d_i-c_x\cdot d_i)/f_x,(y_i\cdot d_i-c_y\cdot d_i)/f_y,d_i].$
After that, the relative velocities in the robot frame can be easily computed as $\Bar{vv}_i=(\pp_{1i}-\pp_{0i})/(t_1-t_0)$. The corresponding absolute velocities $\vv_i$ can be computed from $\Bar{vv}_i$ and the instantaneous velocity of the robot. Moreover, the sizes of moving obstacles appearing in the image can be estimated from the bounding box estimated using MaskRCNN. Lidar data can further compute the radius of obstacles that are outside the camera's FOV.
Compared to tracking-based velocity estimation, our method utilizes pixel-level motion information. Thus, our approach does not suffer from correspondence or loss of tracking target. Moreover, our FlowNet and MaskRCNN can take advantage of pre-trained models, which are trained on large-volume datasets and are able to generalize well to the real world as well as the simulated environments (see Fig. 5).
% $(p1[0]*z - _cx*z)/_fx, (p1[1]*z - _cy*z)/_fy, z$
% \subsection{Sensor Fusion}
% \begin{wrapfigure}[14]{r}{0.4\linewidth}
% % \vspace{-20pt}
%   \begin{center}
%     \includegraphics[width=\linewidth]{fig/sensors.png}
%     % \vspace{-10pt}
%   \end{center}
%   \vspace{-5mm}
%   \caption{The FOV of the camera and the Lidar. The combination of both sensors is used to estimate the state of the obstacles and pedestrians.}
% %   \vspace{-20mm}
%   \label{fig:sensor}
% \end{wrapfigure}
% Although perception by the camera has reasonable accuracy, networks are still black-boxes, and cameras usually have a relatively small field of view. In our approach, we also use 2D Lidars to enhance the detection from the camera. The Lidar and camera sensors are complementary, as illustrated in Figure \ref{fig:sensor}. The Lidar provides ditances of obstacles to robot in a wider range, while the camera can provide velocity information of the objects in front of the robot. 

% Moreover, current object detection neural networks automatically neglect static obstacles (e.g. walls) by treating them as background and not segmenting them. Although the deficiency of walls usually has little influence on semantic meanings, it will hugely undermine collision avoidance and navigation. Lidar can provide relatively accurate relative distances to the robot. Therefore, we use Lidar to detect nearby static obstacles, by transforming relative distances and angles to the frame of the robot's coordinate. 

\section{COLLISION AVOIDANCE}

\label{sec:navigation}

% % We then propose a novel approach for solving differential driving issues. 

%\subsection{VO Algorithm with Partial Observations}

\begin{figure}
\centering
\begin{tabular}{@{}c@{\hspace{1mm}}c@{\hspace{1mm}}c@{}}
    \includegraphics[width=.4\linewidth,height=.5\linewidth]{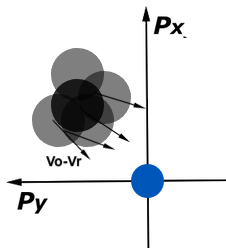} &
    \includegraphics[width=.4\linewidth,height=.5\linewidth]{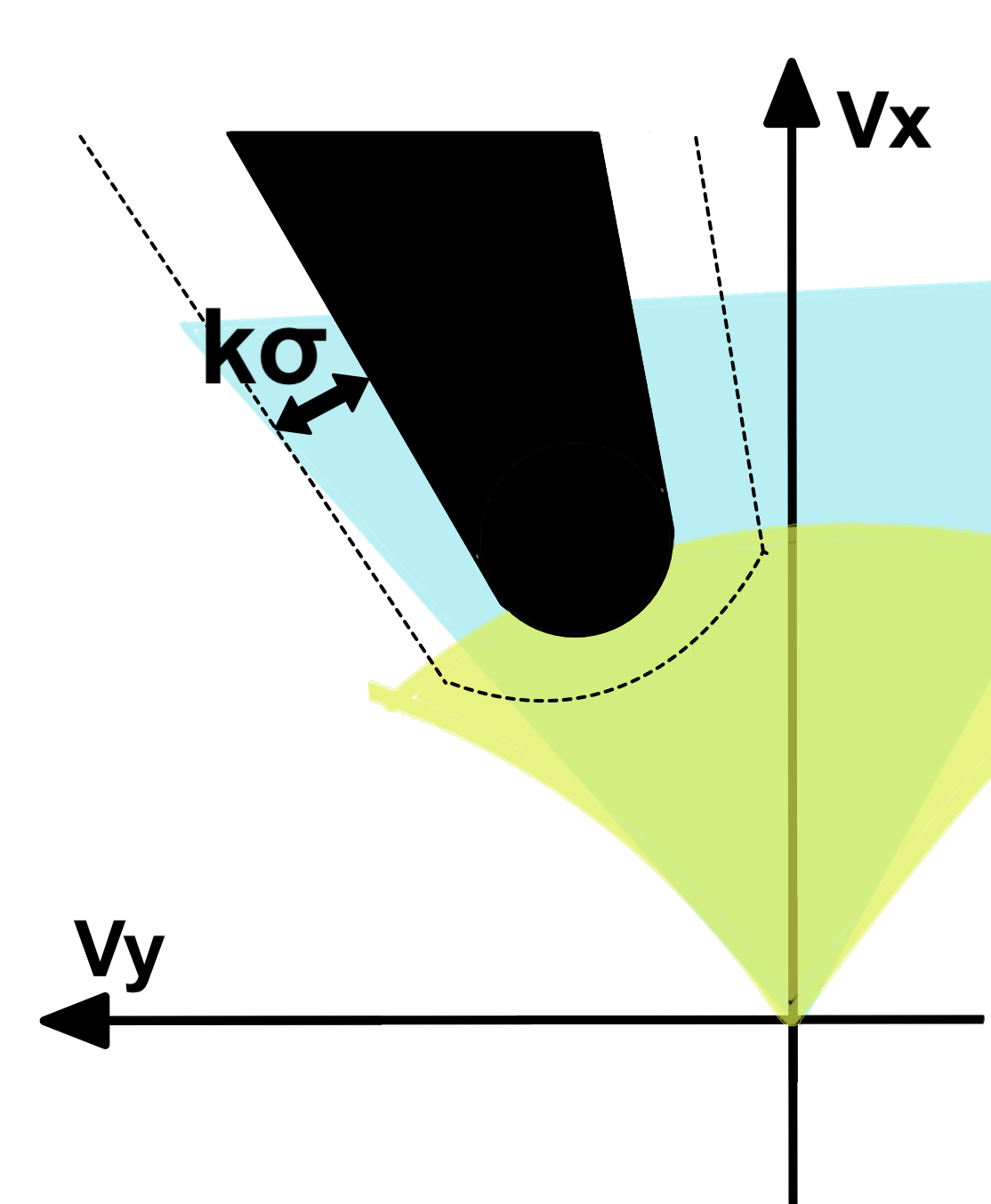} \\
    \small (a) Positions & \small  (b) Velocity Obstacle
\end{tabular}
\caption{
Modified Velocity Obstacles algorithm for partial observation of obstacles. (a): In a given frame, the blue circle is the robot with radius $r_r$ and velocity $\vv_r$. The black area corresponds to one obstacle with the Minkowski sum of radius $r_o+r_r$, with relative velocity $\vv_o-\vv_r$. (b): The velocity obstacle is the black area and its nearby area within distance $k\sigma$, which is used in our probability collision avoidance. We give a bound on the confidence based on the parameter $k$ in Section \ref{sec:Probabilistic}. The blue area is the field-of-view of the camera. The green area is computed based on kinematic constraints of the robot. During each step, the robot has to choose one feasible velocity in the intersection of the blue and green areas, while keeping $k\sigma$ away from the black region.}

\label{fig:VO}
\vspace{-4mm}
\end{figure}

After detecting nearby obstacles using our perception algorithm, 
% based on the Velocity-Obstacle(VO) algorithm, 
our navigation algorithm takes into account the uncertainty of perception, the constraint of partial observation, and kinematic constraints of the  robot.
%AMRCOMMENT: by "property" do you mean "likelihood"?

\subsection{Probabilistic Collision Avoidance}

\label{sec:Probabilistic}

We first account for the sensor errors and derive bounds for probabilistic collision avoidance. To accomplish this goal, we model obstacles' parameters as Gaussian distributions and compute the velocity constraints for the robot. Our formulation uses a parameter $k$ in Lemma \ref{lm:chance_constraint}, which is used to guarantee the confidence in collision avoidance.
%Tin Lemma \ref{lm:chance_constraint}.

Our bound derivation makes some assumptions on the motion of dynamic obstacles, which mostly correspond to pedestrians. We assume that there would be no large or sudden change in the velocity of the pedestrians that are in close proximity to the robot. In addition, the pedestrians will not intentionally run into a robot from behind, even though they are not in the field-of-view of the robot sensors.
% Our bound derivation makes some assumptions on the dynamic obstacles. We assume that a moving agent would slow down if it moves close to the robot. Moreover, a moving agent would prevent collision if it is behind the robot. 

Our approach builds on prior techniques based on reciprocal velocity obstacles~\cite{orca, VO}. Given the position $\pp_o$ and velocity $\vv_o$ of an obstacle with radius $r_o$, let the robot's position be $\pp_r$, velocity $\vv_r$, and radius $r_r$. The velocity obstacle defined as Equation \ref{eqn:VO}.

\begin{equation}
\begin{split}
    V_{r|o}^T = \left \{\mathbf{v}  | \exists  t \in \left [ 0, T \right ] :: t\mathbf{v} \in D\left ( \pp_o-\pp_r, r_r+r_o \right )  \right \}\\
    D(\mathbf{p},r) = \left \{ \mathbf{q} | \left \| \mathbf{q}-\mathbf{p}  \right \|< r \right \},
\end{split}
\label{eqn:VO}
\end{equation}
where D is the circular disk centered at position $\mathbf{p}$ with radius $r$.
Because of the inaccuracies and uncertainties in real-world measurements, we assume the observed obstacles have errors in positions and velocities with different probabilities. In our formulation, the positions and velocities of obstacles are represented by Gaussian distributions, as shown in Figure \ref{fig:VO}, where gray circles are obstacles with potential positions and the arrows correspond to the potential velocities.
% and can be extended to non Gaussian formulations.
We assume each vector is isotropic in all directions, the elements in the same vector are independent and their variances have the same relationship with respect to the distance:
\begin{align}
    \mathbf{p}_i \approx \mathcal{N}_2(\mu_p, \Sigma_p(d))\\
    \mathbf{v}_i \approx \mathcal{N}_2(\mu_v, \Sigma_v(d)), 
    \mathbf{v}_r \approx \mathcal{N}_2(\mu_r, \Sigma_r),
    \label{eqn:Gaussians}
\end{align}
where $\mathbf{p}_i$ is the relative position of the $i_{th}$ obstacle with respect to the robot; $\mathbf{v}_i$ is the velocity of the $i_{th}$ obstacle; d is the distance between the obstacle and the robot, and $\mathbf{v}_r$ is the velocity of the robot. If the obstacle is near the robot, the perception errors in the position and velocity are lower. Based on the formulations, $\Sigma_v(d)=\sigma_v(d)^2\II$ and $\Sigma_p(d)=\sigma_p(d)^2\II$, we observe that $\sigma_v(d)^2$ and $\sigma_p(d)^2$ are monotonically decreasing.

Given a velocity of the robot $\vv$ and  $t\in (0,T]$, lets consider the velocity obstacle formulation given by Equation \ref{eqn:VO}. In this case, we can write the constraint for velocity regions using Equation \ref{eqn:VOConstraints} as:
% \begin{align}
%     f_{i}(\vv,\pp_i,\vv_i) =  \left\| \mathbf{p}_i \right\|^2 - (r_r + r_i)^2 - \frac{((\vv-\vv_i)\trans \pp_i )^2}{\left\| (\vv-\vv_i) \right\|^2} > 0,
%     \label{eqn:VOConstraints}
% \end{align}
\begin{align}
    f_{i}(\vv,t|\pp_i,\vv_i) =  \left\| \mathbf{p}_i - (\vv-\vv_i)t \right\|^2 - (r_r + r_i)^2 \nonumber \\
    f_{i}(\vv, t, \pp_i,\vv_i) > 0,
    \label{eqn:VOConstraints}
\end{align}
where $r_r$ and $r_i$ are the radius of the robot and the obstacle $i$, respectively. $\vv$ and $\vv_i$ represents the velocity of the robot and the $i_{th}$ obstacle at time $t$, respectively. This formulation ensures that the distance between the robot and the obstacle during the next time step is larger than the sum of their individual radius~\cite{manocha1998solving}. We use the following theorem about its distribution. 
% \begin{theorem}
% \label{thm:VO_constraint}
\begin{restatable}[]{theorem}{voconstraint}
\label{thm:VO_constraint}
\it For any velocity $\vv$ of the robot, the distribution of $f_i$, $\dD (f_{i}(\vv|\pp_i,\vv_i))$ 
% can be bound by a Gaussian distribution $\nN (\mu_f(v),\sigma_f(d)^2)$, where
is a non-central $\xX^2$ distribution with different $\sigma$, where the mean is a function of $\vv$ and variance with regards to d is monotonically decreasing. d corresponds to the distance from the obstacle to the robot and is determined when the obstacle is detected.
\end{restatable}

The proof of Theorem \ref{thm:VO_constraint} is given in Section IX   in ~\cite{liang2021ofvo}.
% ~\cite{ARXIV} [CITE THE ARXIV PAPER].
%appendix\footnote{Appendix: \url{https://arxiv.org/abs/2004.10976},\label{appendix}}.
Theorem~\ref{thm:VO_constraint} highlights the property of constraints of the velocity space based on the probabilistic velocity obstacle formulation.

The variables in Equation \ref{eqn:VOConstraints} are all random variables. 
% in order to constraint the probability of collision, with given time step t, we use chance constraint w.r.t the variants in Equation \ref{eqn:VOConstraints} as $P(f_{i}(\vv|\pp_i,\vv_i, t)>0)$. $P(f_{i}(\vv|\pp_i,\vv_i,t)>0) $ is a function of $\vv$, $\pp$ and $\vv_i$, we denote it as $P(\vv|\pp_i,\vv_i,t)$. 
At time step $t$, we use the chance constraint $P(f_{i}(\vv,t|\pp_i,\vv_i)>0)$ to represent the probability of $f_i(\cdot) >0$ in Equation \ref{eqn:VOConstraints}. We denote the function $P(f_{i}(\vv|\pp_i,\vv_i,t)>0) $ as $P(\vv|\pp_i,\vv_i,t)$. Let $\mu_f$ and $\sigma_f$ denote the mean and standard deviation of the distribution given by $f(\cdot)$, respectively. 
For a given $k>0$, we enforce the distribution to have $\mu_f \pm k\sigma_f >0$. Then, $P(\vv|\pp_i,\vv_i,t)$ will increase as $k$ gets larger. The lower bound on the probability is given by Lemma \ref{lm:chance_constraint}:
\begin{restatable}[]{lemma}{chanceconstraint}
    \label{lm:chance_constraint}
    {\it Choose a scalar value $k>0$, and let $\mu_f \pm k\sigma_f >0$, then the chance constraint $P(\vv|\pp_i,\vv_i,t)$ is lower bounded by  $\frac{k^2}{1+k^2}$.}
\end{restatable}

The proof of Lemma \ref{lm:chance_constraint} is given in Section IX  in ~\cite{liang2021ofvo}. In practice, our approach can be rather conservative when $k$ is large.
% . CITE THE PAPER %appendix\footnotemark[\ref{appendix}].
In order to guarantee collision avoidance, for all $\tau \in (0,t]$, the robot's velocity needs to satisfy $P(f_{i}(\vv, \tau |\pp_i,\vv_i)>0)>\frac{k^2}{1+k^2} $. As shown in Figure \ref{fig:VO} (b), if we choose a velocity outside of $k\sigma_f$ distance of the black velocity obstacle area, the confidence of the collision avoidance is bounded by Lemma \ref{lm:chance_constraint}.

In this manner, we can combine the constraints formulated using all nearby obstacles. We use the function $g(\pp_i,\vv_i,t)$ to represent the feasible velocity area related to the obstacle $i$. 
Assume that the set $C_o$ contains all the detected obstacles, the feasible velocities can be computed by applying all the constraints from nearby obstacles, as following equation:
\begin{equation}
    \gG = \bigcap_{i\in C_o}g(\pp_i,\vv_i).
    \label{eqn:all_constraints}
\end{equation}
When we choose a feasible velocity from $\gG$, It is guaranteed (according to Lemma \ref{lm:chance_constraint}) that the confidence of collision avoidance is at least $\frac{k^2}{1+k^2}$.

\subsection{Partial Observations}
% we use a modified Velocity-Obstacle(VO) algorithm for collision avoidance. In this section, we present our collision avoidance algorithm. Moreover, we address the issue of handling partial observations and kinematic constraints of the robot. 

% The velocity obstacle algorithm is widely used for collision avoidance in a dynamic environment. This algorithm calculates relative positions and velocities of nearby obstacles and finds a feasible subset of velocities in which the robot can move. The input of our VO-based algorithm includes probabilistic velocities and positions of obstacles computed from the perception modules, the velocity of the robot and its relative goal location. 

% Although sensors RGB-D can produce information of nearby obstacles, there are still some areas out of camera's FOV and only within the range of the lidar. Moreover, there are areas behind the robot that are not visibile from the camera or the lidar. 
In our model, the largest robot's field-of-view is limited by the field-of-view of the 2D Lidar. 
%The areas behind the robot are not covered because of the assumptions from Section \ref{section:overview} that pedestrians will not intentionally collide with the robot from behind. %With the assumption, we only need to consider the obstacles that are within the range of Lidar and camera. 
For areas inside the range of the camera, the probabilistic VO formulation described above is directly used based on the information about detected obstacles. We define the area of feasible velocities as $V^t_c=\gG$, at time step $t$.

Since the field-of-view of the camera is typically smaller than that of the Lidar, dynamic obstacles may appear in the detection area of this Lidar but not in the field-of-view of the RGB camera. As a result, the velocities of such obstacles are unknown and there could be potential collision during the time step t. We use additional constraints to prevent such collisions. 
Our approach considers two types of dynamic obstacles: robot-like obstacles and pedestrian obstacles. The robot-like obstacles have the same configurations as the robot, and pedestrian obstacles correspond to moving pedestrians, other dynamic objects, etc.
%AMRCOMMENT: FOV i assume is field of view, but this acronym is not previously defined

%Because of the limited FOV of the camera, dynamic obstacles may appear in the detection of Lidar but not of RGB-D camera, then velocities of those obstacles are unknown and there will be a potential collision in timestep T. In order to solve this issue, we add some velocity constraints to the original VO algorithm so that the robot will not collide even in a partial observation situation. We categorize dynamic obstacles into two types, robotic-obstacles with the same configurations as the robot and conscious-obstacles are pedestrians, pets or other dynamic objects.

For robot-like obstacles, if we can ensure that the robot's position during the next step will lie within the field-of-view of its camera, the robots will not collide with each other since they will avoid collisions with the obstacles within their FOV of camera. This constraint is expressed as:  
%  it follows that the robots will not collide with robotic obstacles within the detection area of the camera. As a result, if we choose the velocity within the field of view of the camera, then the robots will not collide with each other. This can be expressed as:
\begin{equation}
      V_c =  \left \{(v_x, v_y) | \left |\arctan(v_y,v_x)\right|< \frac{FOV_c}{2} \right \},
    \label{eqn:vo-partial}
\end{equation}{}
where $FOV_c$ is the field-of-view of the camera and $(v_x, v_y)$ are allowable velocities  of the robot.

For pedestrian-obstacles we assume that there will be no sudden changes in the velocity near the robot at position $\pp_o$. Let the velocity of the obstacle be $\vv_o$. We set the magnitude of velocity $\norm{(\vv_o)} = 1.5m/s$, which is the average speed of pedestrians in normal circumstances~\cite{Levine1999ThePO}. 
This value can change in different conditions. In order to avoid collisions, the robot needs to maintain a sufficient distance from each obstacle during the next time step. Considering the pedestrians which are out of the field-of-view of the robot's camera, the allowable velocity space is given by:
%defined as Equation \ref{eqn:VO_conscious}.} 
% A velocity $\vv_r=(x_r, y_r)$ having a norm of $ \norm{\vv_{0}}=1.5m/s$ is used in our formulation in the Equation \ref{eqn:VO_conscious}.
\begin{equation}
    V_{l}^t=\{\vv| \norm{(\vv-\vv_o) * \tau - \pp_o} > C \}, for \; all \; \tau \in (0,t],
\label{eqn:VO_conscious}
\end{equation}
% where $\vv_2$ is the velocity of the conscious-obstacle. 
where $C$ is the threshold of collision (i.e. the minimum distance between the obstacle and the robot) and $t$ is the time step.

% In many cases, the obstacles have the same maximal speed as the robot. Figure \ref{fig:VO}(a) shows the relative positions and velocities, where $\pp_3$ is the position of the obstacle. %With the constraint of Equation \ref{eqn:vo-conscious}, we have $V_{r|o2}$. 
% Using the similar derivation of, we can compute the constraints of $V_{r|o2}$ for the relative velocities.

% For robotic-obstacles with the same configurations, if we can keep the robot's movement within its current field of view of the camera, which is the blue cone area, robots will never collide with each other, because in the FOV of camera robots know others velocity and can guarantee not to collide. In this case, we set the constraint as Equation \ref{eqn:vo-partial}, where $FOV_c$ is the range of the camera.  In our system, $FOV_c = 70^\circ$. With this constraint, robots can only move within the FOV of the camera, and we can guarantee the robot will not collide with any obstacles with the same configuration as the robot. This constraint limits the robot's velocity in the blue cone area in Figure \ref{fig:VO}(b), and we denote the set of velocities as $V_c$.
% \begin{equation}
%       V_c =  \left \{(v_x, v_y) | \left |\arctan(\frac{v_x}{v_y})\right|< \frac{FOV_c}{2} \right \}.
%     \label{eqn:vo-partial}
% \end{equation}{}

Adding the constraints from partially-observed obstacles, the feasible velocities space is
\begin{equation}
    V_p^t = V^t_c \bigcap V^t_l.
    \label{eqn:multiple_obstacles}
\end{equation}

When calculating $V^t_l$, the potential velocity obstacles are conservative and have larger areas. As a result, $V^t_l$ is smaller than that computed  using the actual sizes of obstacles. If the field-of-view of the camera is larger, we can detect more potential obstacles and their velocities. In this case, $V^t_l$ would be larger and $V^t_p$ would also be larger.
% This formulation of velocity computation assumes an accurate location of the obstacles. In Section VI, we present a probabilistic formulation that takes into sensor errors.
%So far the algorithm is based on accurate detection of obstacles, however perception would always be inaccurate in detecting positions or velocities. In the following section, we analyze the inaccuracy of detection and give the bound of collision avoidance.

\subsection{Kinematic Constraints}

The output of our velocity obstacle algorithm is computed using the Euclidean coordinates. In order to satisfy the kinematic constraints of the robot,
we project the kinematic constraints in the Euclidean space, which is shown as the green area in Figure \ref{fig:VO}(b).
Let $V^t_k$ denote the area computed using the kinematic constraints. As a result, the overall constraints can be represented as $V^t = V^t_p \bigcap V^t_k$ .
At each step, the robot has a preferred velocity of $\vv_\textrm{pref}$, which is calculated according to its current position and goal position~\cite{orca}. In our approach, we choose the best velocity, which is nearest to the preferred velocity from the feasible velocity space, $V^t$. As shown in Equation \ref{eqn:loss_function}, we choose the velocity with the minimum $L$.
\begin{align}
     L = \argmin_{\vv}\left\| \vv-\vv_\textrm{pref} \right\|,\;\; \vv \in  V^t.
     \label{eqn:loss_function}
\end{align}

\section{Benchmark and Results}
The velocity estimation is more accurate for nearby obstacles, and the inaccuracy in position estimation increases based on the distance from the robot. more details are provided in appendix X of the paper \cite{liang2021ofvo}.
% We analyze the errors in our perception module and describe our design choice to compensate the errors in Section X in the appendix of \cite{liang2021ofvo}.
%in the appendix of this full version\footnote{Appendix: \url{https://arxiv.org/abs/2004.10976}}.
In this section, we describe the implementation details and highlight the performance of our approach in real-world and simulated scenarios. We also highlight the improvement in the performance of our approach in scenarios with partial observations. 
%Finally, we compare our method with prior model-based (DWA)~\cite{dwa} and deep reinforcement learning (DRL)~\cite{fan2018fully} methods. 
% In this section, we highlight the implementation details. We also compare performances using different sensors and show the improvement of the performance of our approach in scenarios with partial observation. Finally, we compare our method with other traditional (DWA)~\cite{dwa} and deep reinforcement learning (DRL)~\cite{fan2018fully} methods. 

\subsection{Experimental Setup}
% In the implementation, we use Turtlebot2 as the robot platform, an Astra RGB camera and Hokuyo Lidar. 
% The resolution of the RGB camera is $640\times 480$, and $520$ range data for Lidar. 

In the implementations, we use Turtlebot2 as the robot platform with one camera and one Lidar. We use Astra Orbecc camera with 70-degree field-of-view and $640\times 480$ resolution. The Lidar is Hokuyo 2D Lidar with 240-degree field-of-view and $512$ samples.
We also use Pozyx to localize target relative position to the robot. For computation we use a laptop with an Intel i7-9750H CPU (2.6 GHz) with 32 GB memory and an Nvidia GeForce RTX 2070 GPU. We run our system using Ubuntu 18.04 and ROS Melodic. The segmentation network (MaskRCNN) corresponds to mask-RCNN from detectron2~\cite{wu2019detectron2}, and the optical flow (OFNet) network corresponds to FlowNet2~\cite{ilg2017flownet}.

The underlying simulator used to evaluate our results is Gazebo 9.0.0, with models of Jackal and Turtlebot robots. 
In the simulated environments, we compare the performance DRL, DWA, and PRVO with our approach. DRL is trained by the scenarios described in \cite{fan2018fully}, and DWA implementation is from the ROS package, $dwa \_ local \_ planner$. For these two approaches, the inputs correspond to Lidar data. PRVO implementation is based on \cite{PRVO}. Since PRVO and DWA use many parameters, we tuned them to achieve the highest success rate for each scenario. Since PRVO does not have its perception module, we use ours described in Section IV. PRVO does not consider partial observations, so it treats obstacles in the partially observed area (i.e., out of the field-of-view of the camera) as static obstacles.

\begin{figure*}[t]
  \centering
    \begin{tabular}{@{}c@{\hspace{.3mm}}c@{\hspace{.3mm}}c@{\hspace{.3mm}}c@{}}
  \includegraphics[width=.19\linewidth, height=2.7cm]{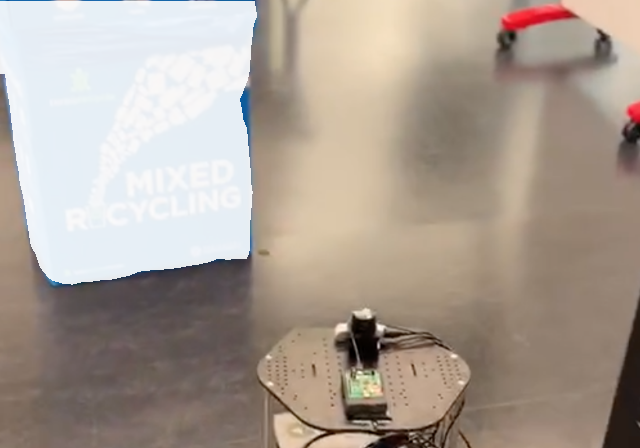} &
  \includegraphics[width=.19\linewidth, height=2.7cm]{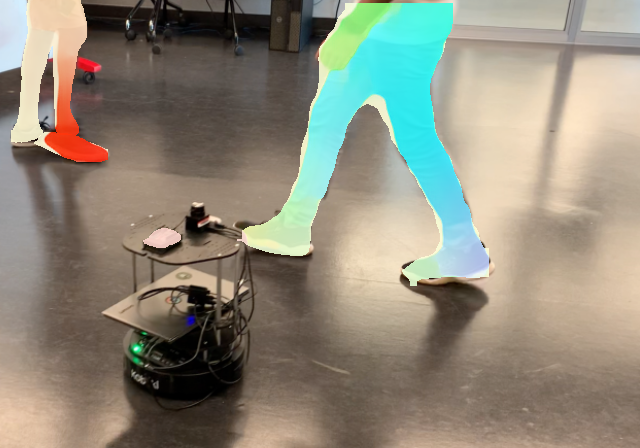} &
  \includegraphics[width=.19\linewidth, height=2.7cm]{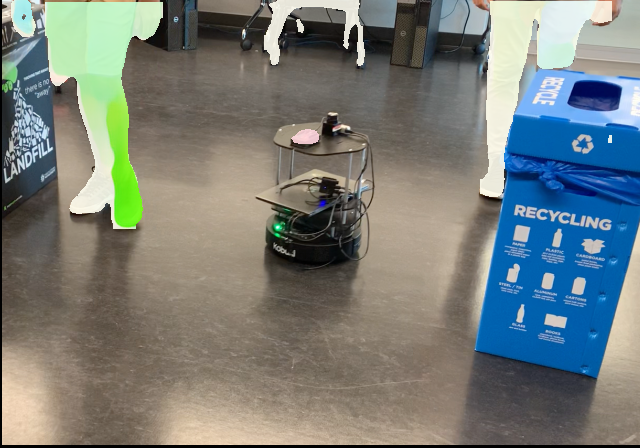} &
  \includegraphics[width=.19\linewidth, height=2.7cm]{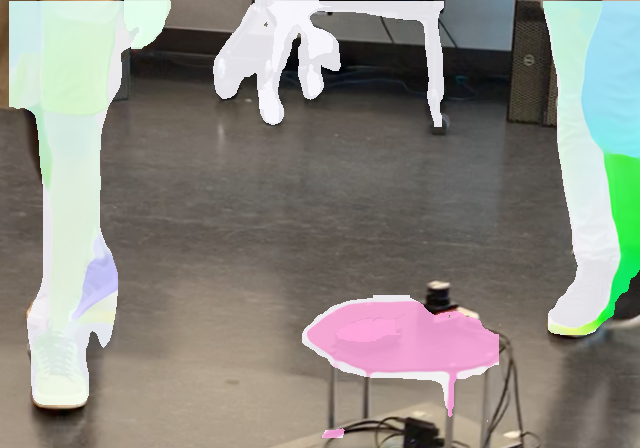} \\ 
  \small (a) Static & \small  (b) Dynamic & \small  (c) Complex & \small  (d) Cross
  \end{tabular}
  \caption {
  Navigation across a room with different configurations: Our algorithm running on Turtlebot2 is used to navigate the robot and avoid collisions with the pedestrians and obstacles. Our algorithm can compute collision-free trajectories for the robot in these scenarios with static obstacles and multiple moving pedestrians. }
  \label{fig:impl}
\end{figure*}

\begin{figure*}[t]
  \centering
    \begin{tabular}{@{}c@{\hspace{.2mm}}c@{\hspace{.2mm}}c@{\hspace{.2mm}}c@{\hspace{.2mm}}c@{\hspace{.2mm}}c@{}}
  \includegraphics[width=.165\linewidth, height=2.7cm]{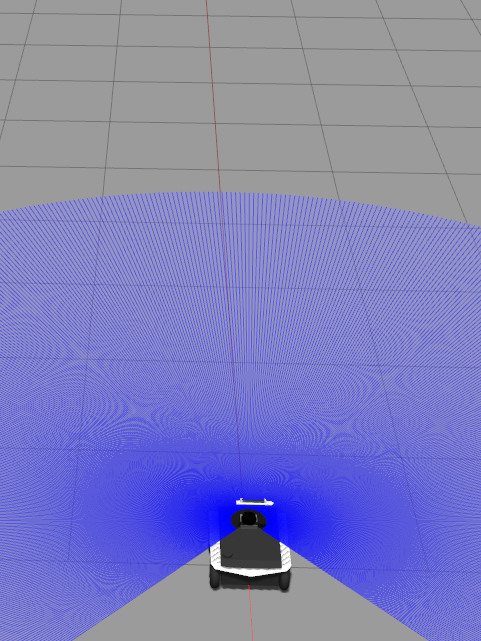} &
  \includegraphics[width=.165\linewidth, height=2.7cm]{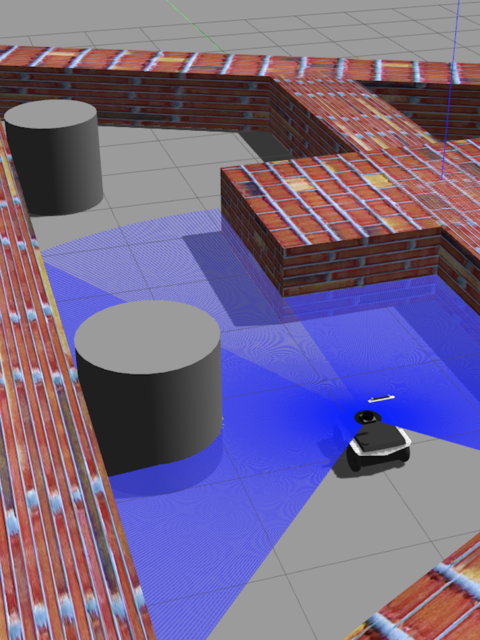} &
  \includegraphics[width=.165\linewidth, height=2.7cm]{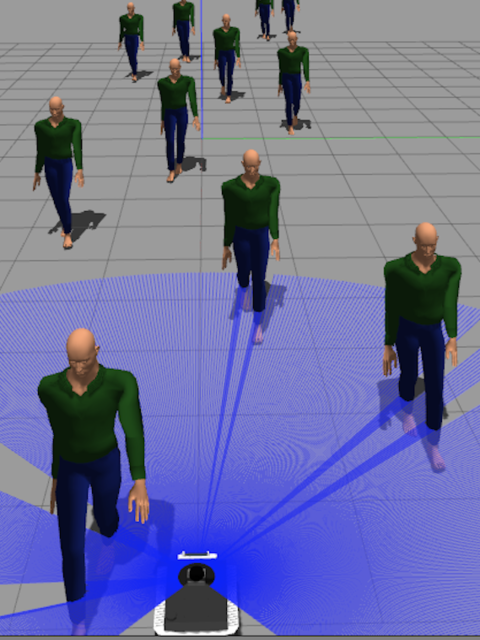} &
  \includegraphics[width=.165\linewidth, height=2.7cm]{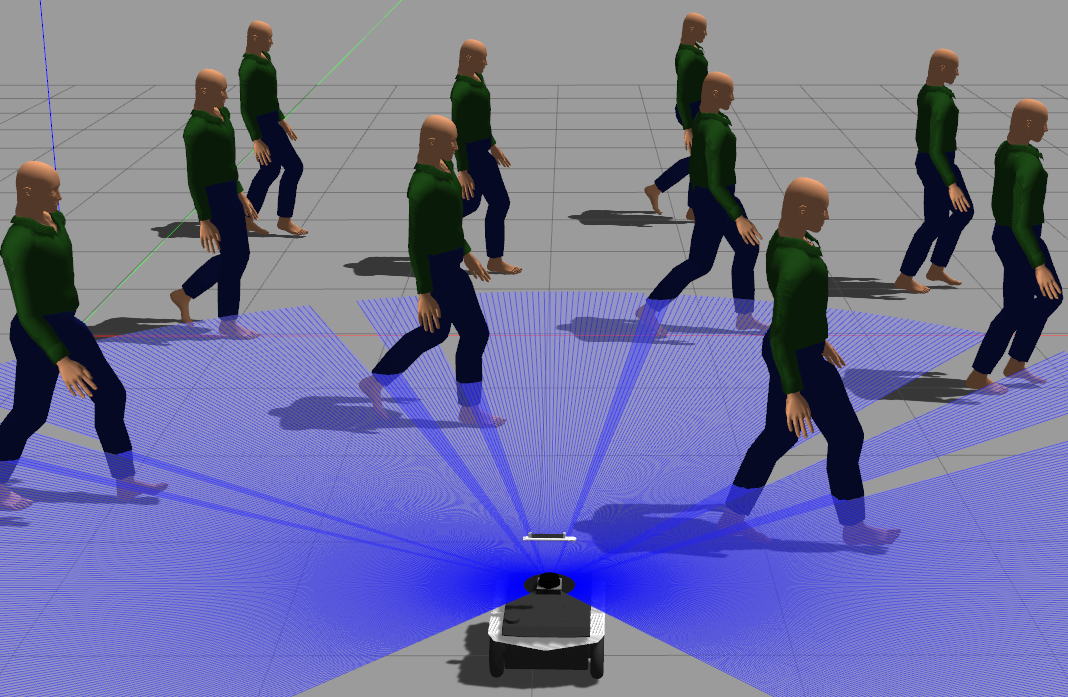} & 
  \includegraphics[width=.165\linewidth, height=2.7cm]{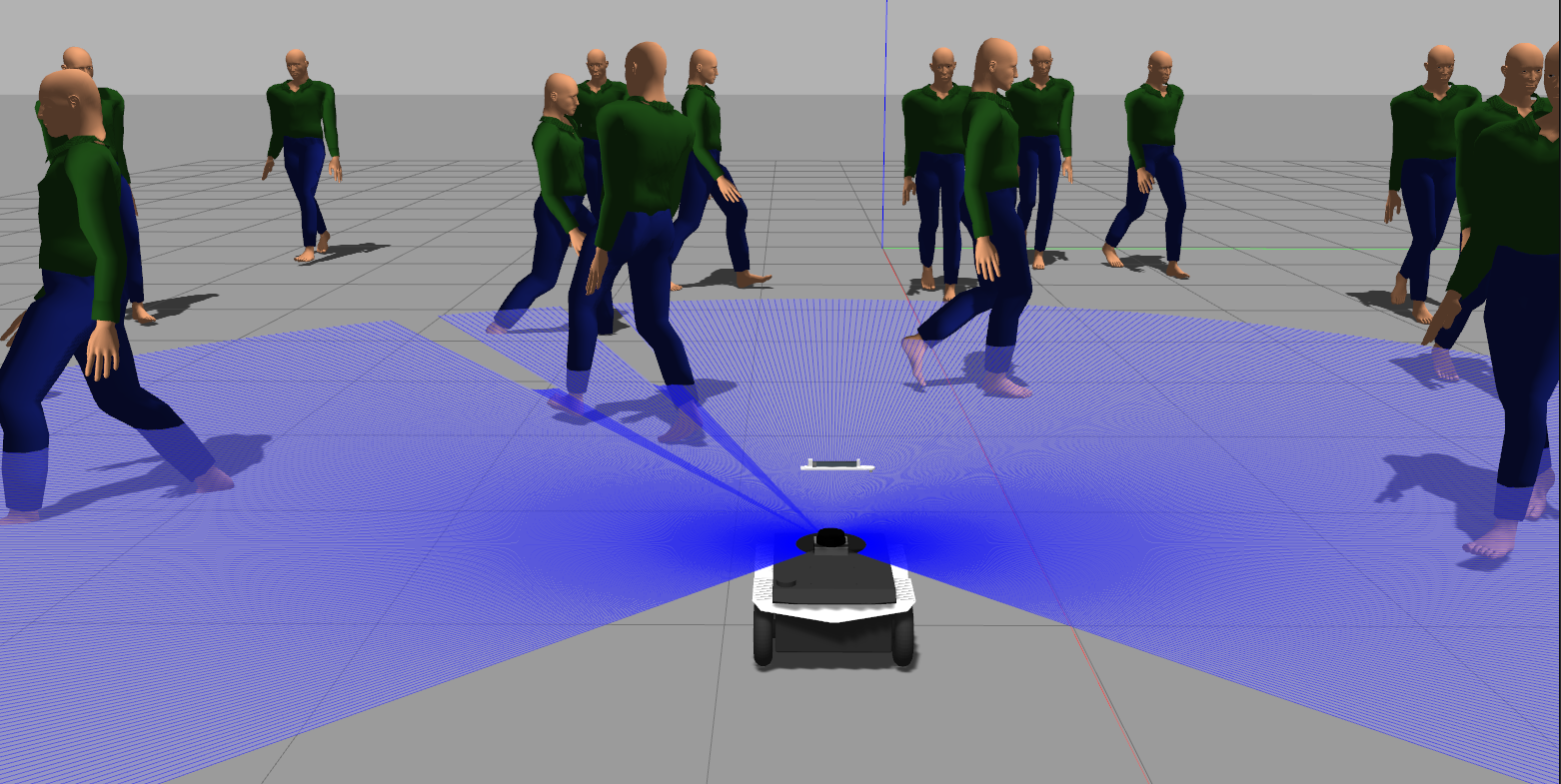} 
  \\ 
%   \small (a) Empty &\small (b) Static & \small  (c) Dynamic & \small  (d) Complex & \small  (e) Cross & \small (f) Social Force
\small (a) Empty &\small (b) Static & \small  (c) Dynamic & \small  (d) Cross & \small (e) Social Force
  \end{tabular}
  \caption {Simulation Scenarios: We use six scenarios to evaluate our algorithm and analyze the performance. The scenes corresponding to (c) and (d) are dynamic scenes with multiple pedestrians. In scenario (e), pedestrians follow the social behavior~\cite{bera2017sociosense}. The robot moves from a start point to a goal point while avoiding static and dynamic obstacles.}
  \label{fig:sim_scene}
%   \vspace{-5mm}
\end{figure*}

% \begin{table}[!htb]
% \centering
% \vspace{-3mm}
% \ra{1.1}
% % \resizebox{1\linewidth}{!}{
%     \small
%     \begin{tabular}{@{}l@{\hspace{6mm}}r@{\hspace{4mm}}r@{\hspace{4mm}}r@{\hspace{4mm}}r@{}}
%         \toprule
%         Success Rate & Lidar+RGB-D & RGB-D  &  Lidar\\
%         \midrule
%         Empty               & xx$\%$  & xx$\%$ &  xx$\%$  \\
% %         \midrule
%         Static Obstacles    & xx$\%$  &  xx$\%$ &  xx$\%$  \\
% %         \midrule
%         Dynamic Obstacles   & xx$\%$  &  xx$\%$ & xx$\%$   \\
% %         \midrule
% Cross Scenario   & xx$\%$  &  xx$\%$ & xx$\%$   \\
% %         \midrule
%         Complex Obstacles   & xx$\%$  & xx$\%$ &  xx$\%$  \\
%         \bottomrule
%     \end{tabular}
% \caption{Success rate using different sensors. Our sensor fusion of RGB-D and lidar achieves higher success rate than only using an RGB-D camera or a Lidar. RGB-D camera can enhance the velocity estimation of dynamic objects, while the lidar helps to avoid static obstacles.
% }
% % }
% \vspace{-3mm}
% \label{tab:ab-sensor}
% \vspace{-5mm}
% \end{table}

\subsection{Simulated and Real-world Scenarios}

% PyTorch 1.4. The segmentation network (MaskRCNN) is mask-RCNN from detectron2~\cite{wu2019detectron2} trained on the COCO dataset~\cite{lin2014microsoft} and the optical flow (OFNet) network is FlowNet2~\cite{ilg2017flownet} trained on FlyingChairs~\cite{DFIB15} and FlyingThings3D~\cite{MIFDB16}. These datasets contain a large amount of real and synthetic data, which enables our networks to generalize well to the data captured in simulated and real-world environments. 

% (e) is the cross scenario. There are static obstacles as well as pedestrians. Our method can successfully drive the robot from the start point to the target while avoiding collisions.

In the simulation, we create different scenarios (as shown in Figure \ref{fig:sim_scene}) to evaluate the performance of our approach and compare with other methods, in Table \ref{tab:quantative}.

1. \textbf{Empty} (Figure \ref{fig:sim_scene}(a)): In this scenario, OF-VO is tested with random goals and start positions. It is used to test if the robot has the basic functionality to go to the goal position.

2. \textbf{Static} (Figure \ref{fig:sim_scene}(b)): This scenario requires the robot to go around static obstacles and reach the goal position.

3. \textbf{Dynamic} (Figure \ref{fig:sim_scene}(c)): In this scenario, there is a crowd of pedestrians moving towards the robot from the front. Pedestrians do not intentionally avoid robots. Instead, the robot needs to avoid the pedestrians and find its way to the goal position.

% 4. \textbf{Complex} (Figure \ref{fig:sim_scene}(d)): In this scenario, there are static and dynamic obstacles. A robot must be able to avoid both types of obstacles.  

4. \textbf{Cross} (Figure \ref{fig:sim_scene}(d)): In this scenario, pedestrians move towards the robot from the sides. This scenario requires the robot to avoid dynamic obstacles from both sides and deal with the issues of partial observation and uncertainties.

5. \textbf{Social Behavior} (Figure \ref{fig:sim_scene}(e)): In this scenario, pedestrians tend to follow social constraints~\cite{bera2017sociosense}. This scenario evaluates the ability of the robot avoiding pedestrians that exhibit human-like social behaviors.

In order to fill the gap between sim-to-real we tested the approach in the real world. Figure~\ref{fig:impl} shows the implementations in four scenarios, where pedestrians are (a) standing still, (b) moving towards the robot, (c) moving between obstacles, and (d) moving in a perpendicular direction.

% \subsection{Ablation Study on Sensors}
% We conduct an ablation study of the sensor choices in Table~\ref{tab:ab-sensor}. We run testing experiments to get the success rate of each model for four types of scenarios, Figure ~\ref{fig:sim_scene}(a to d).

% The results are shown in Table \ref{tab:ab-sensor}. In the empty scenarios, all three models work well and send the robot towards the goal smoothly, as shown in   Figure \ref{fig:sim_scene}(a). In the scenario with only big static obstacles (Figure~\ref{fig:sim_scene}(b)), only the robot with only the RGB-D camera can barely go through the maze. The reason is that the segmentation neural network would neglect walls as background, instead of treating it as an obstacle. Compared with the camera, the Lidar is better at detecting static obstacles.

% In the scene with only dynamic obstacles (Figure \ref{fig:sim_scene}(c)), our perception module can successfully extract the velocities of pedestrians and prominently outperforms  the Lidar-only robot shown in the third row of Table \ref{tab:ab-sensor}, which cannot obtain the obstacle velocities. 

% In the fourth row of Table~\ref{tab:ab-sensor}, there are mixed obstacles corresponding to moving pedestrians and static walls, as shown in Figure \ref{fig:sim_scene}(d). Because Lidar has a big field of view and high accuracy in terms of distance detection, it is used to enhance the position detection of dynamic obstacles by using depth images. As a result, our sensor fusion of RGB-D and Lidar achieves a much higher success rate than either RGBD-only or Lidar-only perception. 

\subsection{Results of Our Approach}
In Table~\ref{tab:quantative} and Table~\ref{tab:confidences}, we highlight the performance that is generated from 200 runs in each scenario. 

\subsubsection{Results of partial-observation}
Because of the limitation of sensors, we have partial observations of the obstacles around the robot. In order to evaluate the performance of different navigation algorithms, we use Scenario \ref{fig:sim_scene}(d) to evaluate collision avoidance performance from partially observed obstacles. In this scenario, pedestrians walk towards the robot from the left and the right sides of the robot.

As table \ref{tab:quantative} shows, our approach has the best performance in avoiding collisions from outside of the view of its camera. 
% In order to compare the performance of our new method with the VO based probabilistic method~\cite{PRVO}, we evaluated the performance of PRVO in scenarios with pedestrians crossing from outside the view of the camera, as shown in Figure \ref{fig:sim_scene}(e).
% DO YOU SHOW THESE RESULTS IN THE PAPER OR VIDEO? WHAT CONCLUSION CAN YOU MAKE ABOUT PRVO IN THIS CASE
% We also perform $200$ runs using each of the algorithms in the crossing scenario in Figure \ref{fig:sim_scene}(e). 
 The success rate using our approach with $k=1$ is 80\%. However, for PRVO the success rate is around 60\%. According to Table \ref{tab:quantative}, DWA does not work well in the crossing scenario. DRL is better because it uses Lidar data for perception, but cannot avoid  all obstacles. These results highlight the benefits of our approach.

% Because of the inaccuracies in perception, the detected positions and velocities of obstacles around the robot are treated as Gaussian distributions as mentioned in section \ref{section:collision_avoidance}. 
% The available velocity area of the robot in each step is a combination of non-Gaussian distributions. 
% Our algorithm chooses the best velocity from the combination of distributions with different constraints. In order to compare the performance of different confidences in collision avoidance, we tested the robot with different confidence parameters in different scenarios.
The performance of our algorithm improves by $33\%$ in the crossing scene, where pedestrians may walk towards the robot from outside the field-of-view of the camera.

\begin{table*}
\newcolumntype{Z}{S[table-format=2.2,table-auto-round]}
\centering
\setlength{\tabcolsep}{3mm}
\ra{1.05}
\small
\begin{adjustbox}{max width=\textwidth}
\begin{tabular}{@{}lccccccccccccccc@{}}
  \toprule
  \multirow{2}[3]{*}{Scenarios} && 
  \multicolumn{4}{c}{Trajectory length (m)}&& 
  \multicolumn{4}{c}{Navigation Time (sec)}&&  
  \multicolumn{4}{c}{Success Rate (\%)}\\
  \cmidrule(l{3mm}r{3mm}){3-6} 
  \cmidrule(l{3mm}r{3mm}){8-11} 
  \cmidrule(l{3mm}r{3mm}){13-16}
   && {OF-VO} & {DWA} & {DRL} & PRVO && {OF-VO} & {DWA} & {DRL} & PRVO &&  {OF-VO} & {DWA} & {DRL} & PRVO \\
  \midrule
  Empty             && 11.3 &\textbf{10.7} &11.1&11.2  &&11.7&11.6&\textbf{11.5}&11.8  &&\textbf{100}&100& 100&100\\ 
  Static obstacles  && 11.4 &12.2&\textbf{10.9}&11.75  &&13.08&14.6&\textbf{11.31}&14.28  &&\textbf{100}&95 & 90&100\\
  Dynamic obstacles && 33.2 &24.1&\textbf{20.2}&39.5  &&65.56&42.7&\textbf{38.6}&74.3  &&\textbf{90} &20 & 55&71\\
  Cross Scenario    && 10.9 &N/A  &\textbf{8.3} &13.8  &&14.93 & N/A &\textbf{10.0}&19.26  &&\textbf{80} &N/A& 20 & 60\\ 
  Social Behavior   && 12.33 &11.9 &\textbf{10.6}&14.7  &&18.1& 14.2&\textbf{12.3}&25.22  &&\textbf{92} &25 & 70&79\\
  \bottomrule
\end{tabular}

\end{adjustbox}
\caption{ We compare OF-VO with DWA~\cite{dwa} and DRL-based algorithm~\cite{fan2018fully} %in terms of accuracy (trajectory length: lower is better), speed (navigation time: lower is better) and reliability (success rate: higher is better). 
in terms of accuracy, speed, and reliability in the simulated scenarios (Fig. \ref{fig:sim_scene}). The number and movement of pedestrians is generated using different models, including social behavior~\cite{bera2017sociosense} in the last row.
%Our method, OF-VO, outperforms other methods in all these scenarios. When the complexity of the environment increases, our method exhibits better performance than the other two. DWA has the shortest length in the empty scenario because it takes into consideration accelerations. Our method achieves a much higher success rate in dynamic scenes and is more reliable. This highlights the benefits of our hybrid approach.
OF-VO has the highest success rate, especially in highly dynamic environments that include dynamic obstacles and crossing scenarios. DWA and DRL result in collisions with the pedestrians, when the pedestrians' density and velocities increase. 
%There is no available data for DWA in cross scenarios, because it fails in this benchmark. Since our method tends to take more conservative strategies to avoid collisions, the robot would take more time and result in a longer 
%trajectory. 
}
\vspace{-6mm}
% \caption{Performance of Ablation Study: }
\label{tab:quantative}
\end{table*}

\subsubsection{Results of Collision Avoidance}
We compare our algorithm (OF-VO) with two other collision avoidance methods, the Dynamic-Window Approach (DWA) and a Deep Reinforcement Learning-based (DRL) approach\cite{fan2018fully}. We compare these three algorithms using three criteria:  

\textbf{Trajectory Length}: This value gives information about the performance of each algorithm; the shortest length indicates better accuracy in terms of planning and perception.

\textbf{Navigation Time}: This value shows how long each algorithm takes to navigate the robot towards the goal. It gives information about the processing speed and the time taken by the navigation algorithm.

\textbf{Success Rate}: This value shows the percentage of the successful cases in 200 runs without collision in each scenario.

Results in Table~\ref{tab:quantative} indicate that our method has the highest success rate in terms of avoiding collisions. DWA and DRL exhibit similar performance in the static scenario. In dynamic scenarios, DWA and DRL are more likely to collide with the pedestrians. However, the DRL method performs better than DWA because DRL also considers dynamic obstacles in training. Because of the advantages of learning based algorithms, DRL exhibits better optimality in terms of running time.  As compared with DWA and DRL, our method tends to be more conservative in terms of avoiding collisions.
We highlight the differences between our approach and PRVO~\cite{PRVO} in Section VIII of \cite{liang2021ofvo}. Compared with PRVO~\cite{PRVO}, our method has a higher success rate, smaller trajectory length, and takes less time in terms of reaching the goal position. PRVO doesn't work well in in dynamic and cross scenarios (Fig. \ref{fig:sim_scene}(d)). As the complexity of the scenario increases in terms of number of pedestrians, the difference in the trajectory length (and time) between our approach and PRVO increases. Overall, PRVO is  more conservative than our approach, and we observe better navigation performance using our approach.

Table~\ref{tab:confidences} shows that with a higher value of $k$, the success rate is higher because the confidence of collision avoidance is higher. However, with a higher value of $k$, the trajectory length is longer, because the robot tends to be more conservative and tries to maintain a larger distance from obstacles.

%outperforms the other two algorithms in terms of trajectory length, navigation time, and success rate. Our method (OF-VO) can reach the goal in the shortest distance and time because that our model-based VO can find a path to the target without redundant oscillations or freezing. This highlights the reliability of our approach

%As the table shows, in empty scenarios, all the models of DRL, DWA and OF-VO have good performance in terms of reaching the goal. Since DWA is better in terms of velocity optimization with acceleration control, the trajectory of DWA is the shortest. However, our OF-VO is the fastest approach. For scenarios containing both static and dynamic obstacles, our method outperforms the two methods in three areas. The table also shows that DWA is less stable in scenarios with dynamic obstacles. In these scenarios, if an obstacle runs across the robot from its side, DWA cannot avoid the collisions. This highlights the improved performance and reliability of OF-VO.

\begin{table}
\centering
\begin{tabular}{lcccc}
  \toprule
  Success Rate(\%)/time(s) & k=0.1 & k=0.7    & k=1 & k=2 \\
  \midrule
  Dynamic obstacles     & 60/22.90 & 76/45.67 & 90/65.56 & 97/101.40\\ 
  Cross obstacles       & 16/10.1 & 64/11.51  & 80/14.93 & 83/23.67\\ 
  Social Behavior & 57/10.13& 79/16.62        & 92/18.1& 96/33.93\\
  \bottomrule
\end{tabular}
\caption{Success rate and running time with different values of $k$ in our approach. The table shows that when the confidence of collision avoidance is higher, the successful rate is higher. However, with higher confidence of collision avoidance, the navigation algorithm tends to be more conservative, and this results in longer time to reach the goal.}
% YOU DON'T SHOW THE TIME TO REACH A GOAL. ALSO MENTION THAT THE PERFORMANCE IN THE TOP 3 ROWS IS FROM YOUR OF-VO ALGORITHN, WHILE YOU USE THE SOCIAL FORCE MODEL (CITATION) IN THE LAST ROW FOR WHICH BENCHMARK. EXPLAIN BETTER SUCCESS RATE WITH YOUR METHOD>
\label{tab:confidences}
%The higher confidence we set, the better performance we can get in collision avoidance. Confidence does not affect static obstacles much because they are relatively accurately detected and has no velocities. However scenarios with more dynamic obstacles, confidence matters for robot to avoid them.

\vspace{-11mm}
\end{table}
\section{CONCLUSIONS AND LIMITATIONS}
In this paper, we present a hybrid navigation algorithm that combines learning-based perception and model-based collision avoidance. We have implemented OF-VO on a Turtlebot with commodity visual sensors, including an RGB camera and a Lidar.  We have evaluated the performance in complex dynamic scenes with multiple pedestrians and highlighted the benefits over prior model-based (DWA) and learning-based (DRL) methods in terms of success rate and reliability.
%Compared to alternative reinforcement learning model, robot is driven by our algorithm can reach the goal in shorter time with higher success rate. 

Although our method performs well in our benchmarks, it has some limitations. In particular, the Velocity-Obstacle is a local navigation method, and the robot is constrained so that it does not move backward. As a result, the robot may get stuck in an impasse, where there is no feasible velocity  (i.e. freezing behavior). Moreover, the optical flow estimation has lower accuracy for large displacements in the obstacles. We currently model the sensor errors using a simple Gaussian mixture model formulation, though it can be extended to other non-Gaussian noise distributions~\cite{park2020efficient}. Another limitation is that our formulation can be conservative or could result in collisions if there are fast-moving obstacles outside the field-of-view of the camera. Our approach directly computes the speed of the robot without considering much about the continuity of the velocities, which leads to some jerky motions in real-world cases. As part of future work, we would like to overcome these limitations and combine our work with global navigation methods. We would also like to evaluate the performance in complex and outdoor scenes with varying number of pedestrians and relax the assumptions on pedestrian motion. In real world scenarios the pedestrian movement is governed by various constraints, including social and cultural factors, and we would like to handle such scenarios.
%Therefore, possible future works are 1) to cooperate method with global navigation method; 2) to increase the frame rate, which can narrow the displacement between consecutive frames and consequently the improve the velocity estimation.

% \section{ACKNOWLEDGEMENTS}
% This work was supported in part by ARO Grants W911NF1910069 and W911NF1910315, Semiconductor Research Corporation (SRC), and Intel.

%%%%%%%%%%%%%%%%%%%%%%%%%%%%%%%%%%%%%%%%%%%%%%%%%%%%%%%%%%%%%%%%%%%%%%%%%%%%%%%%

% \clearpage
% \begin{appendices}
% \label{sec:appendix}
% \input{appendix.tex}
% \end{appendices}

{\small
\bibliographystyle{IEEEtran}
\bibliography{paper}
}

\newpage
\centerline{\LARGE \bf Appendix}
% \begin{theorem}
% \label{thm:VO_constraint}
% For any velocity $\vv$ of the robot, the probability of $f_i$, $Prob(f_{i}(\vv|\pp_i,\vv_i))$ 
% % can be bound by a Gaussian distribution $\nN (\mu_f(v),\sigma_f(d)^2)$, where
% is a noncentral $\xX^2$ distribution with different $\sigma$ where the mean is a function of $\vv$ and variance with regards to d is monotonically decreasing, where d is the distance from the obstacle to the robot, and is determined when obstacle is detected. 
% \end{theorem}

\section{Compare with PRVO}
This section talks about the differences between PRVO \cite{PRVO} and our approach in collision avoidance. 

Our algorithm considers partial observation but PRVO does not, so when pedestrians go to two sides of the robot, PRVO is in risk to collision. 

\begin{figure}
\centering
\begin{tabular}{@{}c@{\hspace{1mm}}c@{\hspace{1mm}}c@{}}
    \includegraphics[width=.42\linewidth]{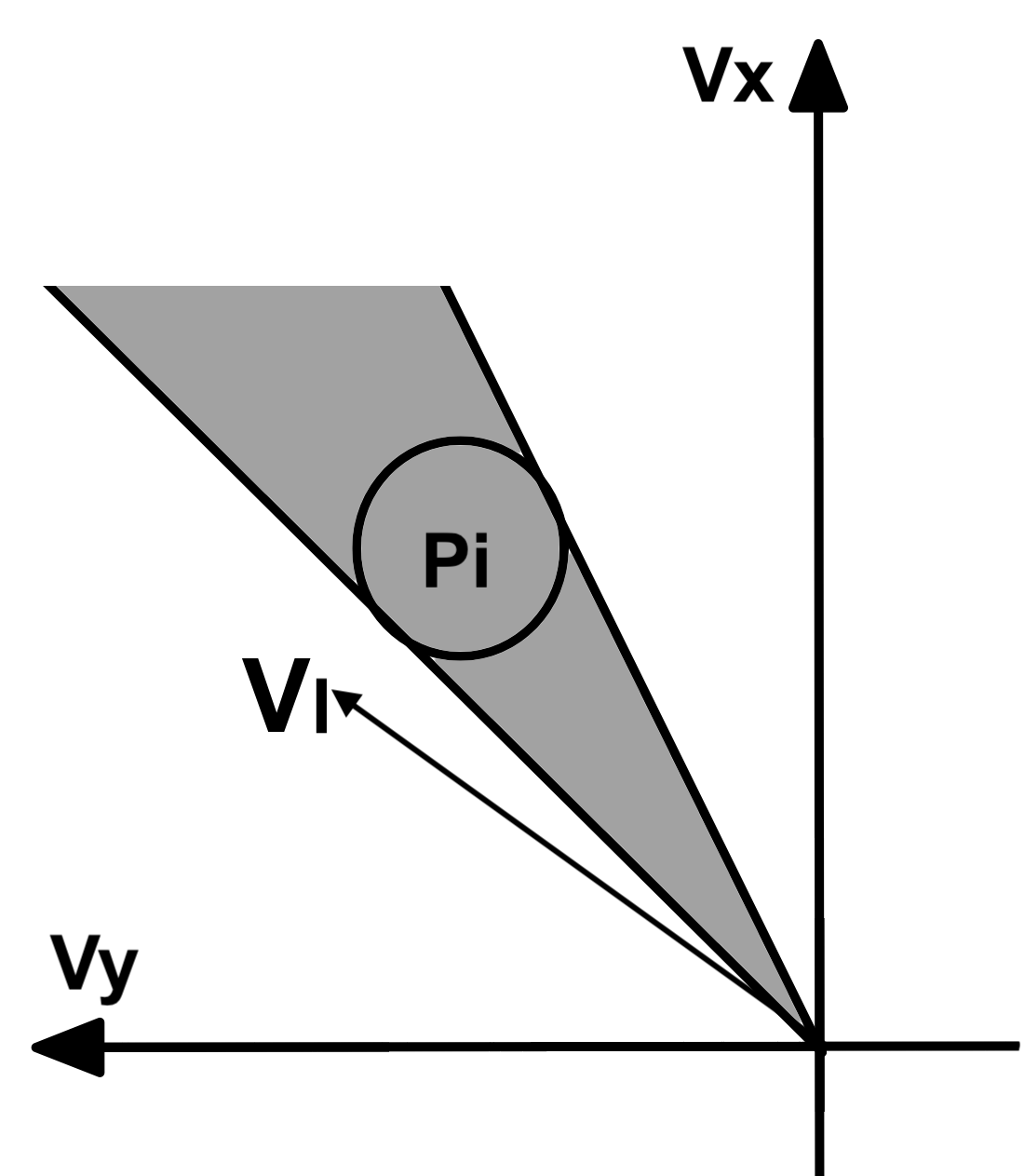} &
    \includegraphics[width=.42\linewidth]{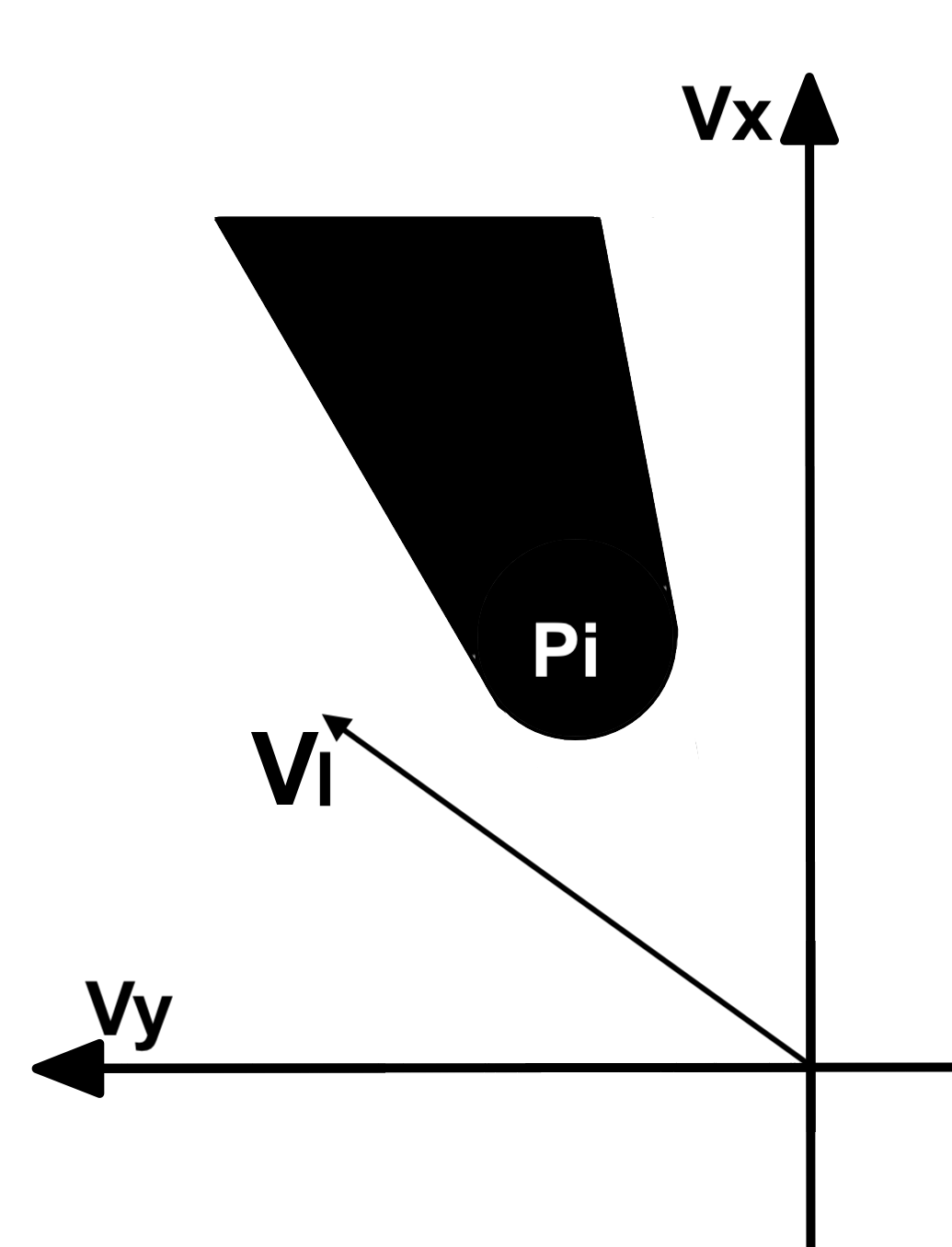} \\
    \small (a) PRVO & \small  (b) OF-VO
\end{tabular}
\caption{In these two figures, $v_l$ is the relative velocity of robot regarding to obstacle. $p_i$ represents the obstacle. The velocity obstacle in our formulation is shown in the black region in (b). In contrast, the velocity obstacle space in PRVO's formulation~\cite{PRVO} corresponds the gray space in (a). The black area is smaller than the grey one, and this means there more open space for available velocities.}
\label{fig:compare_PVO}
\end{figure}

As Figure \ref{fig:compare_PVO} shows, the black area in Figure \ref{fig:compare_PVO} (b) is our velocity obstacle area, and the gray area in Figure \ref{fig:compare_PVO} (a) is the velocity obstacle by PRVO. Our approach gives more accurate and smaller area of velocity obstacle of one object. Thus, there are less open space for PRVO to choose available velocities, so the PRVO is more conservative. The computation of the velocity obstacle spaces in our algorithm is given in Theorem \ref{thm:VO_constraint}.

\section{Theorem and Lemma}
\voconstraint*
% \begin{theorem}
% \label{thm:VO_constraint}
% For any velocity $\vv$ of the robot, the distribution of $f_i$, $\dD (f_{i}(\vv|\pp_i,\vv_i))$ 
% % can be bound by a Gaussian distribution $\nN (\mu_f(v),\sigma_f(d)^2)$, where
% is a non-central $\xX^2$ distribution with different $\sigma$, where the mean is a function of $\vv$ and variance with regards to d is monotonically decreasing, where d is the distance from the obstacle to the robot and is determined when the obstacle is detected.
% \end{theorem}

\begin{proof}
Denote $p_i = (x_i^p,y_i^p)$ with mean $(\mu_x^p, \mu_y^p)$ and variance $\sigma_p(d)^2\II$; $v_i = (x_i^v,y_i^v)$  with mean $(\mu_x^v, \mu_y^v)$ and variance $\sigma_v(d)^2\II$ and $\vv=(v_r^x,v_r^y)$. Because $\pp_i$ and $\vv_i$ are all isotropic, we have $\dd_\textrm{rel}(x,y)$ as the distribution of calculated points with mean $\mu_\textrm{rel}$ and variance $\Sigma_\textrm{rel}(d)$:
\begin{align}
    \dd_\textrm{rel}(x,y) = \pp_i- \vv  t + \vv_i t \approx \nN (\mu_\textrm{rel},\Sigma_\textrm{rel}(d)) \\
    \mu_\textrm{rel} = (\mu_\textrm{rel}^x,\mu_\textrm{rel}^y)=( \mu_x^p+\mu_x^v t-v_r^x t,\mu_y^p+\mu_y^v t-v_r^y t) \\
    \Sigma_\textrm{rel}(d) = (\sigma_p(d)^2+\sigma_v(d)^2t^2)\II
\end{align}
According to Noncentral chi-squared distribution, the distribution of $l_2$ norm of $d_\textrm{rel}$ is represented as $\xX_2(v|\pp_i,\vv_i)$, and d is the detected distance of the obstacle:
\begin{align}
    \mu_{norm} = 2+(\mu_\textrm{rel}^x)^2+(\mu_\textrm{rel}^y)^2=k(v) \\
    \sigma_{norm} = 4(\sigma_p(d)^2+\sigma_v(d)^2t^2)(1+(\mu_\textrm{rel}^x)^2+(\mu_\textrm{rel}^y)^2)=g(d)
    \label{eqn:norm_function}
\end{align}
Then we have $\mu_f(\vv)=\mu_{norm}-(r_r+r_i)^2$ and $\sigma_f(d)=g(d)$ as mean and standard deviation of $f_i$.
According to \ref{eqn:norm_function}, since $\sigma_p(d)^2$ and $\sigma_v(d)^2$ are all monotonically decreasing w.r.t. d, the variance of $f_i$ is also monotonically decreasing.
%TODO: try to prove that its lower bound could be gaussian
%considering the first  \cite{}, we model lower bound of distribution $f_i(\vv,\pp_i,\vv_i)$ as a Gaussian distribution $\nN(\mu_f(v), \sigma_f)$.
\end{proof}

% \begin{lemma}
%     \label{lm:chance_constraint}
%     Given a scalar value $k>0$, the chance constraint $P(\vv|\pp_i,\vv_i,t)$ is bounded by  $\frac{4}{9k^2}$
% \end{lemma}
% \begin{lemma}
%     \label{lm:chance_constraint}
%     { Given a scalar value $k>0$, and letting $\mu_f \pm k\sigma_f >0$, then the chance constraint $P(\vv|\pp_i,\vv_i,t)$ is bounded by  $\frac{k^2}{1+k^2}$.}
% \end{lemma}
\chanceconstraint*
\begin{proof}
    In order to satisfy the constraint function $P(f_{i}(\vv|\pp_i,\vv_i,t)>0)$, we can give a bound to function $f_i$ to constraint the random variable to be positive, then we have:
    \begin{align}
        (X-\mu_f) - k|\sigma_f| > 0
    \end{align}
    According to Cantelli’s inequality, we have:
    \begin{align}
        P(f_{i}(\vv|\pp_i,\vv_i,t)>0) > \frac{k^2}{1+k^2}
    \end{align}
\end{proof}

\section{Error Analysis of Perception Networks}

% and use that to formulate our probabilistic collision avoidance algorithm

%Overview: list the points, explain 2, 
%explain details of sensor noise in perception
%explain details of simulator in simulation

% 1. SENSOR NOISE: probabilities
% 2. assumptions and protocal of human behavior
%     human prevent collision from running behind vehicle
%     human would walk slower towarding robot from front 
%     speed of walking
% 3. fidelity of simulator

% Guarantees and error analysis:
% 1. perceptinos
% 2. probability boundings

\begin{figure}
\centering
\begin{tabular}{@{}c@{\hspace{1mm}}c@{\hspace{1mm}}c@{}}
    \includegraphics[width=.5\linewidth]{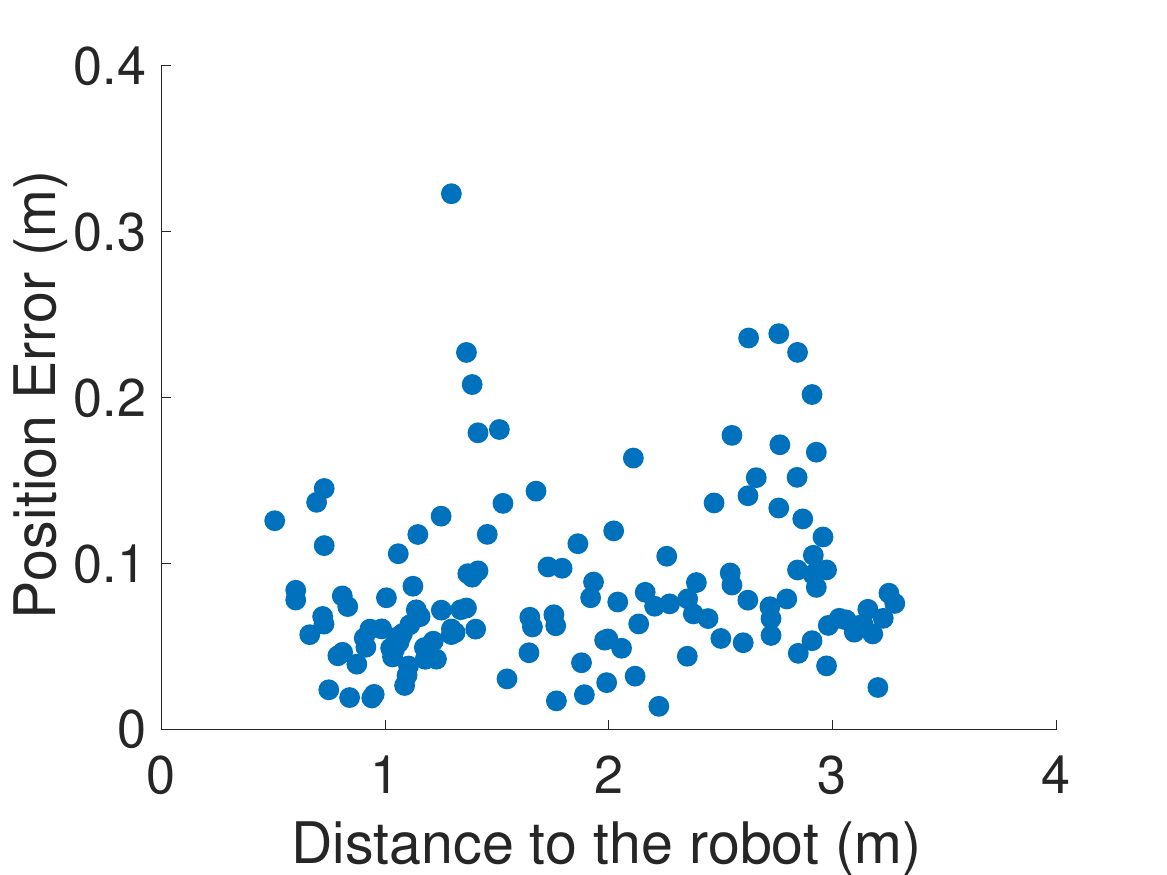} &
    \includegraphics[width=.5\linewidth]{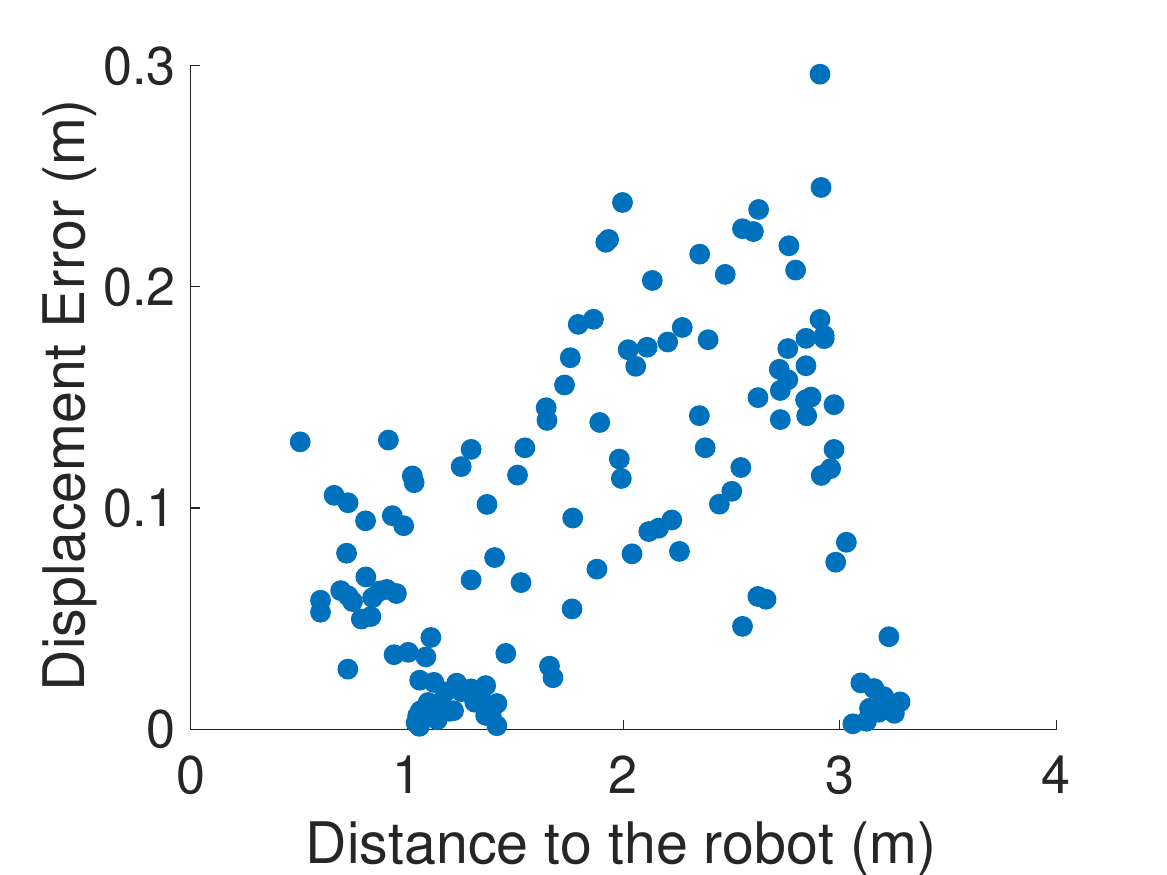} \\
    \small (a) Position Error & \small  (b) Displacement Error
\end{tabular}
% \vspace{-4mm}
\caption{Error in distance and displacement estimation: (a) shows that the errors in position estimation are distributed uniformly along the obstacles' distance. (b) shows that the errors in displacement estimation increase with the distance. The estimation has the highest accuracy at 1 meter distance. }
\vspace{-3mm}
\label{fig:error_perception}
\end{figure}

%Perception would inevitably bring in uncertainty and inaccuracy. 
% In this Section, we analyze these inaccuracies, for collision avoidance. 
%Therefore, we run our algorithm in a simulator and quantify the error in perception, which can be used in the probabilistic navigation method described in the next section. 
In this section, we analyze the errors from our perception algorithm. We estimate the position and displacement of obstacles between consecutive frames. The velocities of obstacles can be computed by the displacement, robot velocities, and time steps. We plot the L2 error of estimations in Figure~\ref{fig:error_perception}.
The results indicate that the displacement (velocity) estimation is more accurate for nearby obstacles, and the inaccuracy in position estimation increases based on the distance from the robot.
%The mean errors are 0.09 m and 0.08 m for position estimation and displacement estimation, respectively.

%Unlike end-to-end deep learning-based collision avoidance methods where it is hard to analyze the performance, our method explicitly uses the output of neural networks (i.e. partial observation) as intermediate results. As a result, it has better interpretability. We can analyze the estimation errors and provide a certain level of guarantee with our approach.

We analyze the accuracy of the two networks used in our perception algorithm. As reported by FlowNet2~\cite{ilg2017flownet}, the average endpoint error in KITTI 2012 is $e_f=4.09$, which means that predicted optical flows are $4.09$ pixels away from the ground truth (on average). 
For an object with depth $z$, an error of $e_f$ pixels in optical flow would result in  an error of $e_f\cdot z/f$ to its displacement estimation. In our implementation, we have focal lengths $f_x=f_y=457$, so the error for an object at $2m$ away would be $0.02m$. 
The error of MaskRCNN is harder to quantify. From Figure~\ref{fig:teaser}, we can see that the segmentation of the pedestrian is reasonably good in our benchmarks. The average precision (.50 IoU) of MaskRCNN is 62.3\%~\cite{he2017mask} on the COCO dataset~\cite{lin2014microsoft}, which means nearly two-thirds of predictions have more than .50 IoU. In order to make our collision avoidance more conservative, we enlarge the bounding box of moving obstacles by more than 50\%. Since the networks we use are pre-trained on large datasets mixed with both simulation and real-world images, the sensor noise is largely covered in the training data. We did not observe a significant difference between the simulation and real performance. %In our experiments where the speed of pedestrians is 1 m/s, the average absolute error is $0.076m$ for position estimation and $0.11m/s$ for velocity estimation. This error is tolerable in benchmarks when we use a slightly larger bounding box for the robot and obstacles.
% To make our method even more robust, Lidar is used to detect the distance from the surrounding obstacles. As a result, the robot is aware of nearby objects even when the camera-based detection fails.
% \textcolor{red}{If we have time, show differences between ground-truth in ROS and detection of networks}
%Based on the analysis of these inaccuracies, we assume the obstacles near the robots have probabilistic positions and velocities.
% (e.g., Gaussian or non-Gaussian).
%We use that information to derive the bounds.
%In Section \ref{section:collision_avoidance} we give an algorithm to handle the probabilistic property of perception.

\end{document}